\documentclass[fleqn,10pt]{wlscirep}
\usepackage[utf8]{inputenc}
\usepackage[T1]{fontenc}

\usepackage{cite}
\usepackage{amsmath,amssymb,amsfonts,bm}
\usepackage{algorithm}
\usepackage{algorithmic}
\usepackage{graphicx}
\usepackage{url}
\usepackage{hyperref}
\usepackage{textcomp}
\usepackage[acronyms]{glossaries}
\usepackage{xcolor}
\usepackage{nomencl}

\usepackage{enumitem}
\def\BibTeX{{\rm B\kern-.05em{\sc i\kern-.025em b}\kern-.08em
    T\kern-.1667em\lower.7ex\hbox{E}\kern-.125emX}}

\usepackage{todonotes}
\usepackage{booktabs}
\usepackage{multicol}
\usepackage{multirow}
\usepackage{svg}
\usepackage{verbatim}
\usepackage{caption}
\usepackage{subcaption}
\usepackage{tabularx}
\usepackage{comment}
\usepackage{rotating}

\newcommand{\cctrl}{c}

\newcommand{\urob}{\bm{u}^{\text{rob}}}
\newcommand{\unomrob}{\bm{u}^{\text{ref}}}
\newcommand{\umin}{\bm{u}^{\min}}
\newcommand{\umax}{\bm{u}^{\max}}
\newcommand{\usm}{u^{\text{smooth}}}
\newcommand{\srob}{\bm{x}^{\text{rob}}}
\newcommand{\smin}{\bm{x}^{\min}}
\newcommand{\smax}{\bm{x}^{\max}}
\newcommand{\rrob}{\bm{r}^{\text{rob}}}
\newcommand{\rhorob}{\rho^{\text{rob}}}
\newcommand{\vrob}{v^{\text{rob}}}
\newcommand{\omrob}{\omega^{\text{rob}}}
\newcommand{\wrob}{w^{\text{rob}}}

\newcommand{\xirob}{\xi^{\text{rob}}}

\newcommand{\Np}{N^{\text{p}}}
\newcommand{\Nc}{N^{\text{c}}}
\newcommand{\rhoobs}{\rho^{\text{obs}}}
\newcommand{\xrob}{x^{\text{rob}}}
\newcommand{\yrob}{y^{\text{rob}}}
\newcommand{\throb}{\theta^{\text{rob}}}
\newcommand{\rstobs}{\bm{r}^{\text{static,obs}}}
\newcommand{\rdynobs}{\bm{r}^{\text{dyn,obs}}}
\newcommand{\wdynobs}{w^{\text{dyn,obs}}}

\newcommand{\xidyn}{\xi^{\text{dyn,obs}}}

\newcommand{\sobs}[1]{\bm{x}^{\text{#1,obs}}}
\newcommand{\xobs}[1]{x^{\text{#1,obs}}}
\newcommand{\yobs}[1]{y^{\text{#1,obs}}}
\newcommand{\vxobs}[1]{v_{x,#1}^{\text{dyn,obs}}}
\newcommand{\vyobs}[1]{v_{y,#1}^{\text{dyn,obs}}}
\newcommand{\xref}{\bm{x}^{\text{ref}}}
\newcommand{\tildex}{\Tilde{\bm{x}}^{\text{rob}}_{\kappa}(\Np)}
\newcommand{\tildeu}{\Tilde{\bm{u}}^{\text{rob}}_{\kappa}(\Nc,\Np)}
\newcommand{\isobs}{\mathcal{I}^{\text{static,obs}}}
\newcommand{\idobs}{\mathcal{I}^{\text{dyn,obs}}}
\newcommand{\rhoplan}{\rho^{\text{plan}}}
\newcommand{\rhosafe}{\rho^{\text{safe}}}
\newcommand{\xat}{x^{\text{att}}}
\newcommand{\yat}{y^{\text{att}}}

\newacronym{apf}{APF}{Artificial Potential Function}
\newacronym{hlrrt}{HL-RRT*}{Horizon-based Lazy Rapidly-exploring Random Tree}
\newacronym{mpc}{MPC}{Model Predictive Control}
\newacronym{poa}{PoA}{Point of Attraction}
\newacronym{rrt}{RRT*}{Rapidly-exploring Random Tree}
\newacronym{sar}{SaR}{Search and Rescue}
\newacronym{uav}{UAV}{Unmanned Aerial Vehicle}
\newacronym{usar}{uSaR}{urban Search and Rescue}
\newacronym{htmpc}{HP+TMPC}{}

\newcommand\norm[1]{\left\lVert#1\right\rVert}

\title{Enabling Robots to Autonomously Search Dynamic Cluttered Post-Disaster Environments}

\author[1]{Karlo Rado}
\author[1,*]{Mirko Baglioni}
\author[1]{Anahita Jamshidnejad}
\affil{Delft University of Technology, Control and Operations Department, 2629 HS, Delft, The Netherlands}

\affil[*]{M.Baglioni@tudelft.nl}


\begin{abstract}
Robots will bring search and rescue (SaR) in disaster response to another level, in case they can 
autonomously take over dangerous SaR tasks from humans. 
A main challenge for autonomous SaR robots 
is to safely navigate in cluttered environments with uncertainties, while avoiding static and moving obstacles.
We propose an integrated control framework for SaR robots in dynamic, uncertain environments, including a computationally efficient heuristic motion planning system that provides a nominal (assuming there are no uncertainties) collision-free trajectory for SaR robots and a robust motion tracking system that steers the robot to track 
this reference trajectory, taking into account the impact of uncertainties. 
The control architecture guarantees a balanced trade-off among various SaR objectives, while handling the hard constraints, including safety. 
The results of various computer-based simulations, presented in this paper, showed significant out-performance (of up to 42.3\%) of the proposed integrated control architecture compared to two commonly used state-of-the-art methods (Rapidly-exploring Random Tree and Artificial Potential Function) in reaching targets (e.g., trapped victims in SaR) safely, collision-free, and in the shortest possible time.
\end{abstract}

\begin{document}

\flushbottom
\maketitle

\thispagestyle{empty}

\begin{table}[!htb]
    \caption{\footnotesize Frequently used mathematical notations}
    \label{tab:math_notations}
    \begin{tabularx}{\textwidth}{|l|l|l|X|}
        \hline
        \multicolumn{1}{|l|}{\footnotesize{\textbf{Notation}}} & \multicolumn{1}{l|}{\footnotesize{\textbf{Explanation}}} & \multicolumn{1}{l|}{\footnotesize{\textbf{Notation}}} & \multicolumn{1}{l|}{\footnotesize{\textbf{Explanation}}}                  
        \\ \hline
        \footnotesize{$\kappa$} & \footnotesize{Discrete time step} & \footnotesize{$\xobs{static}(o)$} & \footnotesize{The $x$ position of static obstacle $o$} \\ 
        \footnotesize{$\cctrl$} & \footnotesize{Control sampling time} & \footnotesize{$\yobs{static}(o)$} & \footnotesize{The $y$ position of static obstacle $o$} \\ 
        \footnotesize{$\Np$} & \footnotesize{Prediction horizon of MPC} &
        \footnotesize{$\xobs{dyn}_{\kappa}(o)$} & \footnotesize{The $x$ position of dynamic obstacle $o$ at time step $\kappa$} \\ 
        \footnotesize{$\xrob_{\kappa}$} & \footnotesize{The $x$ position of the robot at time step $\kappa$} & 
        \footnotesize{$\yobs{dyn}_{\kappa}(o)$} & \footnotesize{The $y$ position of dynamic obstacle $o$ at time step $\kappa$} \\
        \footnotesize{$\yrob_{\kappa}$} & \footnotesize{The $y$ position of the robot at time step $\kappa$}  & \footnotesize{$\vxobs{\kappa}(o)$} & \footnotesize{Horizontal velocity of dynamic obstacle $o$ at time step $\kappa$} \\
        \footnotesize{$\throb_{\kappa}$} & \footnotesize{Heading angle of the robot at time step $\kappa$} & \footnotesize{$\vyobs{\kappa}(o)$} & \footnotesize{Vertical velocity of dynamic obstacle $o$ at time step $\kappa$} \\
        \footnotesize{$\vrob_{\kappa}$} & \footnotesize{Linear velocity of the robot at time step $\kappa$} &
        \footnotesize{$\rhoobs$} & \footnotesize{Radius of the obstacle or of the smallest circular area encountering it} \\
        \footnotesize{$\omrob_{\kappa}$} & \footnotesize{Angular velocity of the robot at time step $\kappa$} &
        \footnotesize{$\xref_{\kappa}$} & \footnotesize{Vector of robot's reference states for time step $\kappa$} \\
        \footnotesize{$\rhorob$} & \footnotesize{Radius of the robots (assuming circular shapes}) &
        \footnotesize{$\unomrob_{\kappa}$} & \footnotesize{Vector of robot's reference control inputs for time step $\kappa$} \\ 
        \hline 
    \end{tabularx}
\end{table}

\section*{Introduction}

Autonomous search and rescue (SaR) robots are emerging in post-disaster response, in order to reduce 
the exposure of human SaR staff to life-threatening risks.  
Structural damages and collapsed buildings reshape an environment post disaster. 
Thus, reliable and effective control methods for autonomous, safe navigation in dynamic and (partially) unknown environments are crucial for SaR robots \cite{murphy2014disaster,LiuNejat,rajan2021disaster,deKoning2023,motionplanning:dynamic,motionplan:realtime}. 
Another main challenge for autonomous navigation of SaR robots is avoiding the obstacles 
that move according to generally nonlinear trajectories, without (significantly) compromising the mission criteria. 
State-of-the-art solutions either rely on assuming static obstacles only, or provide 
ad-hoc solutions that cannot be generalized or provide guarantees on the performance of SaR missions via robots \cite{pandey2017mobile, ohki2010collision, liu2024design}.%

To tackle these challenges, we propose a novel control architecture for autonomous SaR robots that should reach known target positions in dynamic cluttered post-disaster environments: 
We first divide the mission planning of SaR robots into two stages, motion planning and optimal motion tracking.  
We extend the greedy heuristic path planning method\cite{Jamshidnejad} that has been introduced to generate a nominal obstacle-free path towards the targets of the robot, in order to account for dynamic obstacles 
that may appear on the way of the robot. 
Note that by nominal, we are referring to a trajectory that is determined assuming there are no uncertainties or uncontrollable disturbances. 
We then integrate this path planning method with an optimal motion tracking system that closely follows the  
nominal trajectory estimated by the heuristic motion planning system while guaranteeing 
robustness to bounded uncontrollable disturbances.%

The optimal motion tracking system is based on a robust version of 
model predictive control (MPC), called tube-based MPC (TMPC).  
TMPC has proven very effective in steering robots for post-disaster SaR \cite{Surma2024}, 
due to its unique capability in providing robustness to bounded uncertainties, 
balanced trade-offs among competing objectives of SaR (e.g., maximizing the area coverage and minimizing the mission time \cite{deKoning2023}), 
systematic handling of the states and inputs constraints, 
and on-the-go stability guarantees \cite{hoy2015algorithms}.%

\section*{Main contributions and road map of the paper}
The main contributions of this paper include:
\setlist[enumerate,1]{leftmargin=1.5cm}
\begin{enumerate}
    \item 
    Addressing the challenge of autonomous robust motion planning for robots in dynamic cluttered environments, improving their multi-objective optimal performance and guaranteeing the mission safety
    \item 
    Leveraging an efficient heuristic path planning approach to adopt dynamic obstacles 
    and plan, in real-time, a shortest path to given targets, based on predicting the potential movements of the obstacles
    \item 
    Integrating the leveraged path planning approach with TMPC, in a novel architecture, that allows 
    to incorporate the impact of model mismatches, in addition to external disturbances, into the control actions
    \item 
    Performing and analyzing the results of extensive case studies, in comparison with two state-of-the-art methods
\end{enumerate}
In the rest of this paper we provide a background discussion, the problem statement and assumptions, 
our proposed methods, the results of the case study with discussions, conclusions, and 
topics for future work.
Moreover, Table~\ref{tab:math_notations} gives the mathematical notation that is frequently used in the paper.%

\section*{Background discussion}
\label{ch:2bg}

Planning the disaster response via robots depends on the disaster's environment. 
Three operational environments are identified for SaR \cite{sar:review121}, 
urban (involving constrained environments\cite{nagasawa_et_al_2021}, e.g., inside a collapsed building),  
wilderness (involving open-ended environments\cite{hashimoto_et_al_2022}, e.g., a forest),  
and air-sea (involving waters\cite{serra_et_al_2020}, e.g., a sea where vessels accidents or water landing has occurred). 
Ground robots\cite{kruijff_et_al_2012}, flying robots \cite{schedl_et_al_2020}, and underwater robots \cite{fattah2016r3diver} may be used in SaR, 
while for various indoor constrained environments ground robots are preferred \cite{colas_et_al_2013}.
Our focus is on urban SaR via ground robots. 
We address the problem of autonomous, effective mission planning and safe navigation of these robots. 
This yields to a generally nonlinear constrained optimization-based problem with multiple competing objectives, 
e.g., reducing the mission time while increasing the area coverage.%

For SaR robots, it is common to use heuristic methods, especially those based on 
shortest path planners and artificial intelligence, e.g., reaction-based swarming \cite{arnold2020heterogeneous}, fuzzy logic control \cite{san2018intelligent}, and ant colony optimization \cite{loukas2008connecting}. 
The main motivation for using these methods is their computational efficiency, making them suitable 
for on-board deployment in SaR robotics. The main shortcomings of heuristic approaches, however, are their ad-hoc case-specific nature and the lack of performance guarantees. 
With advances in computational power and robotics technology \cite{davids2002urban, bogue2019disaster}, incorporating guaranteed mathematical approaches or novel integrated versions 
of them that provide balanced trade-offs between a high performance and computational efficiency 
has been emerging\cite{stecz2020uav, Jamshidnejad, berger2015innovative,de2019autonomous,deKoning2023, baglioni_jamshidnejad_2024}.%

Model predictive control (MPC) is a mathematics-based, systematic control approach 
that determines a sequence of control inputs by optimizing an objective function within a given prediction window, 
satisfying the state and input constraints. 
MPC is implemented in a rolling horizon fashion, i.e., after determining the control input sequence, 
only the first one is injected into the controlled system and the 
MPC problem is solved for the shifted prediction window at the next time step.
MPC has proven very effective for addressing constrained optimization-based 
problems. MPC has often been used in static SaR environments for tracking a reference trajectory that is assumed to be  
provided by another path planning approach \cite{colas_et_al_2013, berger_lo_2015, hoy_et_al_2012, farrokhsiar_et_al_2013}.%

In connection to path planning for robots, exploration and coverage of the SaR environment are both crucial \cite{nattero_et_al_2014, galceran_carreras_2013}. 
Various approaches for area coverage have been proposed based on random search \cite{brooks2009randomised} or artificial intelligence (especially deep learning, reinforcement learning, fuzzy logic control \cite{san2018intelligent, paez_et_al_2021, niroui_et_al_2019}). 
In this regard, MPC has been used for SaR robots to determine control inputs that maximize a reward 
for visiting new parts of the environment  \cite{mohseni_et_al_2017, ibrahim_et_al_2019,de2019autonomous}. 
Another novel application of MPC in multi-objective SaR via robots is through a bi-level architecture \cite{deKoning2023}, where a supervisory MPC level enhances the 
area coverage by re-distributing SaR robots when they locally decide to visit the same or neighboring parts of the environment. 
This division of local and supervisory decision making provides a balanced trade-off between optimizing the global objectives of the SaR mission and meeting the computational requirements.%

Robust versions of MPC \cite{BemporadMorari}, in particular robust tube-based MPC (TMPC), \cite{langson_chryssochoos_rakovic_mayne_2004} deal with bounded uncertainties, 
e.g., bounded disturbances and perception errors that often occur for SaR robots. 
In TMPC, first the nominal version of the MPC problem is solved where no uncertainties are considered  
and the constraints have been tightened compared to the original problem \cite{liu2019recursive}. 
Constraint tightening implies adjusting the bounds on the states and/or control inputs  
(e.g., by shrinking their admissible sets) to ensure that the controlled system 
always operates within its safe operational limits \cite{hoy2015algorithms}, 
although the operational conditions may be affected by larger disturbances than those considered in the decision making procedure.
During the online implementation of TMPC, an ancillary control input 
minimizes the error between the nominal and actual states \cite{mayne2011tube}. 
While the actual state trajectory may deviate from the nominal one, it always remains within a safe bounded region, called 
the \emph{tube}, where the constraints are guaranteed to always be satisfied. 
The diameter of the cross section of the tube is determined according to the maximum 
difference between the nominal and actual states, considering the worst-case uncertainty scenario.%

Navigating SaR robots is often target-driven\cite{wang_tan_nejat_2024} (i.e., by steering the robot via reference points towards a given target). In this case, learning-based methods, e.g., reinforcement learning, are common\cite{guo2009combination,devo_et_al_2020}.
In static environments, navigating safely towards the targets is usually easier than in dynamic environments, where moving obstacles 
may appear on the way of the robot to these targets. 
Using MPC in order to adapt the motion of the robot 
according to the predicted future trajectories of moving obstacles is a promising solution\cite{sani_et_al_2021}.
In this paper, we introduce a novel control architecture for SaR robots that integrates 
heuristic steering methods with systematic MPC-based methods, for improved computational efficiency and performance safety. 

\section*{Problem statement and assumptions}
\label{ch:prob_statement_and_assumptions}

We consider the control problem of a SaR ground robot that should autonomously navigate a dynamic, cluttered, and (partially) unknown environment to reach a known target point. 
The control problem is formulated within a 2D continuous-space framework in discrete time, with time step variable $\kappa$.
In the mathematical derivations, we consider a 
differential drive\cite{dhaouadi2013dynamic} ground robot with a circular shape of radius $\rhorob$, but the proposed methods are adoptable for different types and shapes of robots and 
post-disaster environments. 
For the sake of simplicity of the mathematical formulations, the static and dynamic obstacles all have a circular shape with a fixed radius $\rhoobs$. 
This is equivalent to considering the smallest circular area that encounters an obstacle an area for the robot to avoid.
The robot has a camera that provides visual information within a circular perception field centered around the robot and perceives the shape, position, and velocity of the obstacles.
In addition to static obstacles (e.g., rubble or stones) there are moving obstacles (e.g., humans or falling debris) in the environment.
The control system should determine per control time step the linear and angular velocities that steer the robot towards its next desired states.%

We consider the following assumptions:

\begin{enumerate} [label=\textbf{A\arabic*}]
    \item  
    \label{ass:perception_field_memory}
    Per time step, the robot has perfect information of its own states and scans and gathers information about the part of its environment that falls within the perception field of the robot's camera. The robot keeps no memory of the past.
    \item \label{assumptions_unc_dist_noise} 
    Environmental unmodeled disturbances (e.g., unevenness of the ground) affect the states of the robot (i.e., the position, which is always given for the robot's center, as well as the velocity).
    \item 
    \label{ass:imperfect_state_perception}
    The perception of the robot about the position of the static obstacles is perfect, but for dynamic obstacles this perception is prone to a bounded error (which may be due to, e.g., the motion blur or the obstacle moving faster than the camera's update rate).
\end{enumerate}

\section*{Bi-level control architecture for autonomous safe searching of dynamic cluttered areas}

\begin{figure}
    \centering
    \includegraphics[width=.68\textwidth]{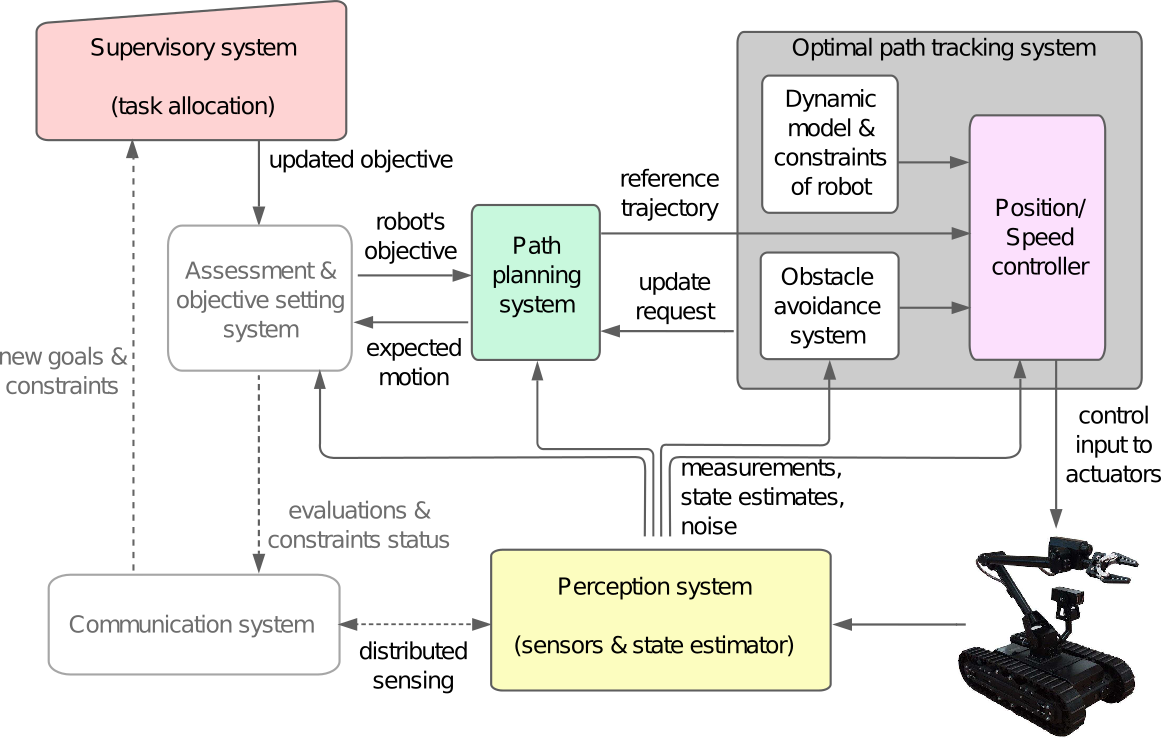}
    \caption{\footnotesize General control architecture proposed for autonomous steering of robots in cluttered dynamic SaR environments.}
    \label{fig:gen_con}
\end{figure}

The control architecture that is proposed for steering a robot in dynamic cluttered SaR environments is illustrated in Figure~\ref{fig:gen_con}: 
In order to improve the computational efficiency and the responsiveness of the steering system of the robot, 
both highly crucial for SaR missions \cite{hierarchicalmpc,basescu2020direct}, 
we opt to separate the steering system 
into a motion planning system and an optimal motion tracking system, which must follow the trajectory 
that is generated by the motion planning system as closely as possible. 
The output of the motion planning system (shown via a green box in Figure~\ref{fig:gen_con}) 
is injected, as the reference trajectory, into the optimal motion tracking system (shown via a gray box in Figure~\ref{fig:gen_con}), 
which will request the motion planning system to re-plan the trajectory whenever needed.%

\subsection*{Motion planning system: Heuristic control}

Due to its ease of implementation and computational efficiency, 
the obstacle-avoiding shortest path approach introduced by 
Jamshidnejad and Frazzoli\cite{Jamshidnejad} was chosen as the basis for the motion planning system. 
There are, however, two main challenges regarding the adoption of this approach 
for a dynamic, cluttered SaR environment that should be addressed: 
First, the approach only includes static obstacles and does not incorporate dynamic ones.
Second, since this approach relies on local knowledge of the robot from its environment, 
its feasibility may be at risk.%

\subsubsection*{Modified obstacle-avoiding shortest path approach for dynamic environments}

\begin{figure}
    \begin{subfigure}{0.32\textwidth}
        \centering
        \includegraphics[width=1\linewidth]{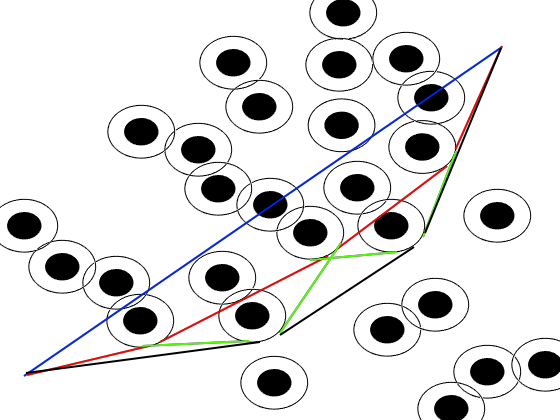}
        \caption{\footnotesize Path smoothening occurs in case of self-intersection: The plot shows the connecting line in blue, the first and final iterations in red and green, and the smoothened path in black.}
        \label{fig:cleaning}
    \end{subfigure}
    \hspace*{1ex}
    \begin{subfigure}{0.32\textwidth}
        \centering
        \includegraphics[width = \linewidth]{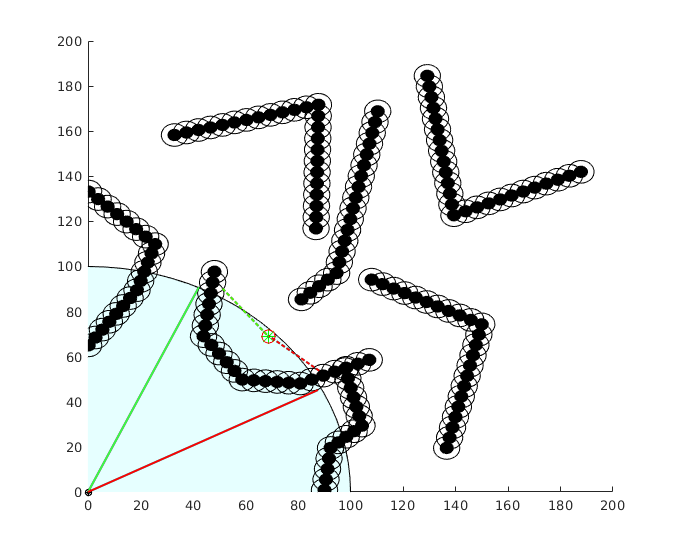}
        \caption{\footnotesize Current candidate paths (solid+dashed red and green), with a target that is unreachable according to current perception, are replaced by  paths with reachable targets (solid green and red).}
        \label{fig:temp_destination}
    \end{subfigure}
    \hspace*{1ex}
    \begin{subfigure}{0.32\textwidth}
        \centering
        \includegraphics[width = \linewidth]{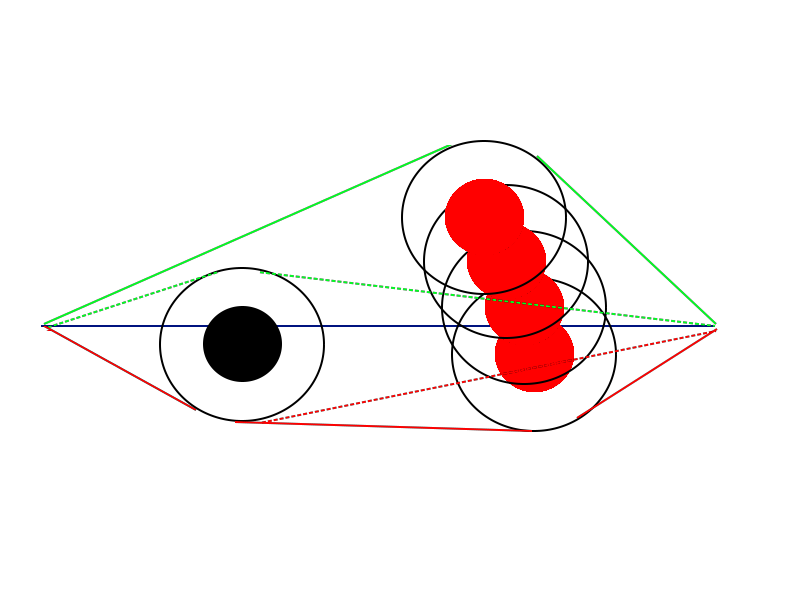}
        \caption{\footnotesize Solid curves are paths that consider all positions of the dynamic obstacle (red circle) within the prediction window. Dashed curves are original paths assuming  
        the obstacle remains static (black circle). }
        \label{fig:obstacle_dynamic}
        \end{subfigure}
    \caption{\footnotesize Main idea behind the heuristic path planning approach\cite{Jamshidnejad}.
    Black hollow circles surrounding static (black) and dynamic (red) obstacles show forbidden areas for the robot.
    In plot \ref{fig:temp_destination} the robot is at the origin and its current perception field   
    is the light blue area.
    In plot \ref{fig:obstacle_dynamic} the robot is at the left-hand side end of the black piece of line and its current target is at the right-hand side end of it. \label{fig:heuristic_path_planning}}
\end{figure}

In the original algorithm\cite{Jamshidnejad}, a greedy heuristic approach has been adopted to avoid obstacles that are composed of any arbitrary union of circular shapes, by generating the path across the tangential arcs to these circular shapes (see Figure~\ref{fig:cleaning}). 
This way various arbitrarily-shaped obstacles can be handled by dividing them  
into smaller pieces and by approximating each piece by the smallest circle that encounters the piece. 
Due to its limited perception field, the robot determines a temporary intermediate target, which it intends to approach, per control time step. The motion planning system evaluates the shortest paths at both sides of the straight path that, 
regardless of the obstacles, connects the current position of the robot and 
the position of its current temporary target.
In case the candidate paths are of the same length, 
one is chosen randomly.
Equidistant points on the shortest path are injected as reference points into the local motion tracking system of the robot. 
In case the resulting shortest path is non-smooth or self-intersecting 
(see the green curves in Figure~\ref{fig:cleaning}), 
a tangent line is drawn across the outer obstacles 
to smoothen the path (see the black curve in Figure~\ref{fig:cleaning}).%

We made the following modifications to the heuristic path planning approach, in order to make it suited for practical cases where the robot should avoid moving obstacles as well and does not store detailed past information for efficiency of the on-board computations and the memory and energy consumption (see Assumption~\ref{ass:perception_field_memory}): 
Due to the limited awareness of the robot about its environment (limited to the perception field of its camera), it may face situations where the temporary target that connects its current position and the final target  
is unreachable according to the robot's environmental awareness (Figure~\ref{fig:temp_destination}).
We address this by replacing the original candidate paths 
with traversable paths within the detection zone of the robot that 
have an endpoint as close as possible to the unreachable temporary target.%
 
Moreover, in order to incorporate the dynamic obstacles into the heuristic motion planning approach,   
the following steps are taken, assuming perfect knowledge about the dynamic equations of the obstacle: 
\begin{enumerate}[label=Step \arabic*]
    \item 
    \label{step1} 
    The obstacle avoiding path planning approach is implemented to generate 
    traversable, safe paths for the robot considering only the obstacles perceived as static. 
    \item
    \label{step2}
    Using the predicted dynamics of the dynamic obstacles within a given prediction window, 
    all relevant dynamic obstacles are modeled, each as a set of static obstacles located at these predicted positions.
    \item 
    \label{step3}
    The reference points of the robot for all the time steps within this prediction window 
    are specified on the shortest traversable and safe path obtained via \ref{step1}, 
    with all positions of the dynamic obstacle 
    for the time  steps within the prediction window included in the picture of the robot as 
    static obstacles.
    \item 
    \label{stepn}
    In case, by comparing the position of the robot and the static obstacles that represent  
    the position of the dynamic obstacle any risks of 
    collision are detected, the illustrated static obstacles for the time steps with a risk 
    of collision will be united to form an obstacle belt. A new collision-free path is then generated.
    \item 
    Repeat \ref{step1}--\ref{stepn} until the shortest traversable and safe path is determined.
\end{enumerate}
Figure~\ref{fig:obstacle_dynamic} illustrates how the shortest traversable and safe path is determined, by first considering 
only the static obstacle (shown via the black circle) and then by modifying the resulting path 
in order to avoid collisions with the static obstacles that represent the dynamic
obstacle for all the time steps within the given prediction window. 
We assume that the robot moves across this path with linear and angular velocities that ensure a trade-off between performance and safety. This, for instance, is obtained by selecting the higher threshold between half of the maximum velocity of the robot and its midpoint velocity.%

Next, we expand the discussions for situations where the dynamics of the moving obstacles is not perfectly known by the robot, thus the robot's predictions about the future positions of the obstacle are prone to errors.%

\subsubsection*{Including dynamic obstacles when perception errors may exist in the motion planning system}

In practice, especially in cluttered SaR environments that are prone to various 
uncontrolled stimuli, the dynamics of the obstacles cannot be perfectly captured via mathematical models. 
Additionally, the perceived position of dynamic obstacles based on the images of the camera is prone to errors 
(Assumption~\ref{ass:imperfect_state_perception}). 
Therefore, the path planning approach must be made robust to these uncertainties. 
This is guaranteed if the robot safely navigates in the environment even when 
maximum uncertainties are realized, i.e., when the forbidden areas (hollow circles around each dynamic obstacle in Figure~\ref{fig:heuristic_path_planning}) are expanded considering the maximum 
modeling or perception error. 
Moreover, the cumulative errors enhance the uncertainty of the predictions. 
This effect is incorporated by increasing the upper value of the errors according to the proximity of the 
prediction to the current time. 
In fact, a larger prediction window allows to incorporate and assess longer-term impacts of current control 
inputs, which increases the chances of a safe and optimal mission and decreases the risk of recursive infeasibility. 
However, in addition to heavier online computations, 
prediction in larger windows leads to larger cumulative errors and thus risks to safety and degrading the performance. 
This further motivates the introduction of a bi-level control architecture 
that generates a reference path that approximately provides safety and 
desirable performance, and that delegates the obstacle avoiding task to a reference tracking control system 
(see Figure~\ref{fig:gen_con}).   
Whenever the local reference tracking control problem becomes 
infeasible or the performance criteria falls under desirable thresholds, 
the motion planning system 
updates the reference path, based on the updated information.
Meanwhile, the optimal motion tracking system keeps the motions of the robot safe and crash-free.%

\subsection*{Optimal motion tracking system: Robust tube-based model predictive control (TMPC)}

Based on assumptions \ref{assumptions_unc_dist_noise} and \ref{ass:imperfect_state_perception}, 
two sources of uncertainties affect the performance of the robot:  
the unmodeled disturbances that deviate the states of the robot from the planned states and the errors in perceiving the position of dynamic obstacles. 
Therefore, we use robust TMPC in motion tracking 
in order to optimally follow the reference trajectory that is determined by the motion planning system, while systematically incorporating all the constraints into the decision making procedure. 
TMPC uses dynamic mathematical models to predict the states of the robot and of the dynamic obstacles in a given prediction window, 
and uses these predictions to optimize the control inputs. 
The state vector of the robot, which encapsulates all the necessary past information for time step $\kappa$ to predict the future states, is given by $\srob_{\kappa}  = [\xrob_{\kappa},\yrob_{\kappa},\throb_{\kappa}]^\top$, which includes, respectively, the 2D position of the center of the robot and the robot's orientation with respect to the horizontal axis. 
The control input vector that steers the motion of the robot for time step $\kappa$
is given by $\urob_{\kappa}  = [\vrob_{\kappa},\omrob_{\kappa}]^\top$, which includes, respectively, 
the linear and the angular velocities of the robot. 
The obstacles are identified by their position vector $\rstobs(o) = [\xobs{static}(o),\yobs{static}(o)]^{\top}$ 
for static obstacle $o$ and by their state vector $\sobs{dyn}_{\kappa} = [\xobs{dyn}_{\kappa}(o),\yobs{dyn}_{\kappa}(o),\vxobs{\kappa}(o),\vyobs{\kappa}(o)]^\top$ per time step $\kappa$ for dynamic obstacle $o$,  
including the 2D position $\rdynobs_{\kappa}(o)$ and the velocity elements of the obstacle, respectively  
(note that due to the circular shape of obstacles, instead of their orientation or angular velocity, the robot perceives the components of the linear velocity). 
The state evolves due to the accelerations $a^{\text{obs}}_{x,\kappa}(o)$ and $a^{\text{obs}}_{y,\kappa}(o)$ by the driving forces of the obstacle. 
Next, we explain how the motion tracking system predicts the states of the robot and of the dynamic obstacles within a given prediction window.%

\subsubsection*{Dynamic prediction models for the motion of the robot and of the dynamic obstacles}
\label{ch:robot_and_obst_model}

The kinematics equations for translational motion of the centroid of a differential drive mobile SaR robot, as well as the orientation of the robot, both used by the motion tracking TMPC system are given by: 
\begin{subequations}
    \label{eq:rob_dynamics}
    \begin{align}
        \label{eq:rob_dynamics_x} 
        &
        \xrob_{\kappa} = \xrob_{\kappa - 1} + 
        \cctrl \Big(\vrob_{\kappa} \cos\left({\throb_{\kappa -1 }}\right) 
        - \cctrl \omrob_{\kappa} \vrob_{\kappa}\sin\left({\throb_{\kappa-1}}\right) \Big) 
        \\
        &
        \label{eq:rob_dynamics_y} 
        \yrob_{\kappa} = \yrob_{\kappa - 1} + 
        \cctrl \Big( \vrob_{\kappa} \sin\left({\throb_{\kappa-1}}\right) 
        + \cctrl \omrob_{\kappa} \vrob_{\kappa}\cos\left({\throb_{\kappa-1}}\right) \Big) 
        \\
        &
        \label{eq:rob_dynamics_theta} 
        \throb_{\kappa} = \throb_{\kappa - 1} + \cctrl \omrob_\kappa 
    \end{align}
\end{subequations}
The state update equations \eqref{eq:rob_dynamics_x}-\eqref{eq:rob_dynamics_theta}
have been discretized in time using sampling time $\cctrl$, 
i.e., the states are updated every $\cctrl$ time units, while during this interval 
their most recently updated values are used. 
In the proposed architecture, both the motion planning system (in \ref{step3}) 
and the optimal  motion tracking system (as a prediction model embedded in MPC) need to model the dynamics of the moving obstacles. 
In case any knowledge exists or is deducible via filters \cite{esteves_et_al_2024} about the pattern of movement of the obstacles, 
the corresponding kinematics equations may be obtained. 
Otherwise, after perceiving the current position and velocity of the obstacle, the robot assumes that the obstacles will follow the same kinematics equations as the robot itself.%

\subsubsection*{Formulating the TMPC problem of the motion tracking system incorporating the impact of uncertainties}

\begin{figure}
    \centering
    \includegraphics[width = .65\linewidth]{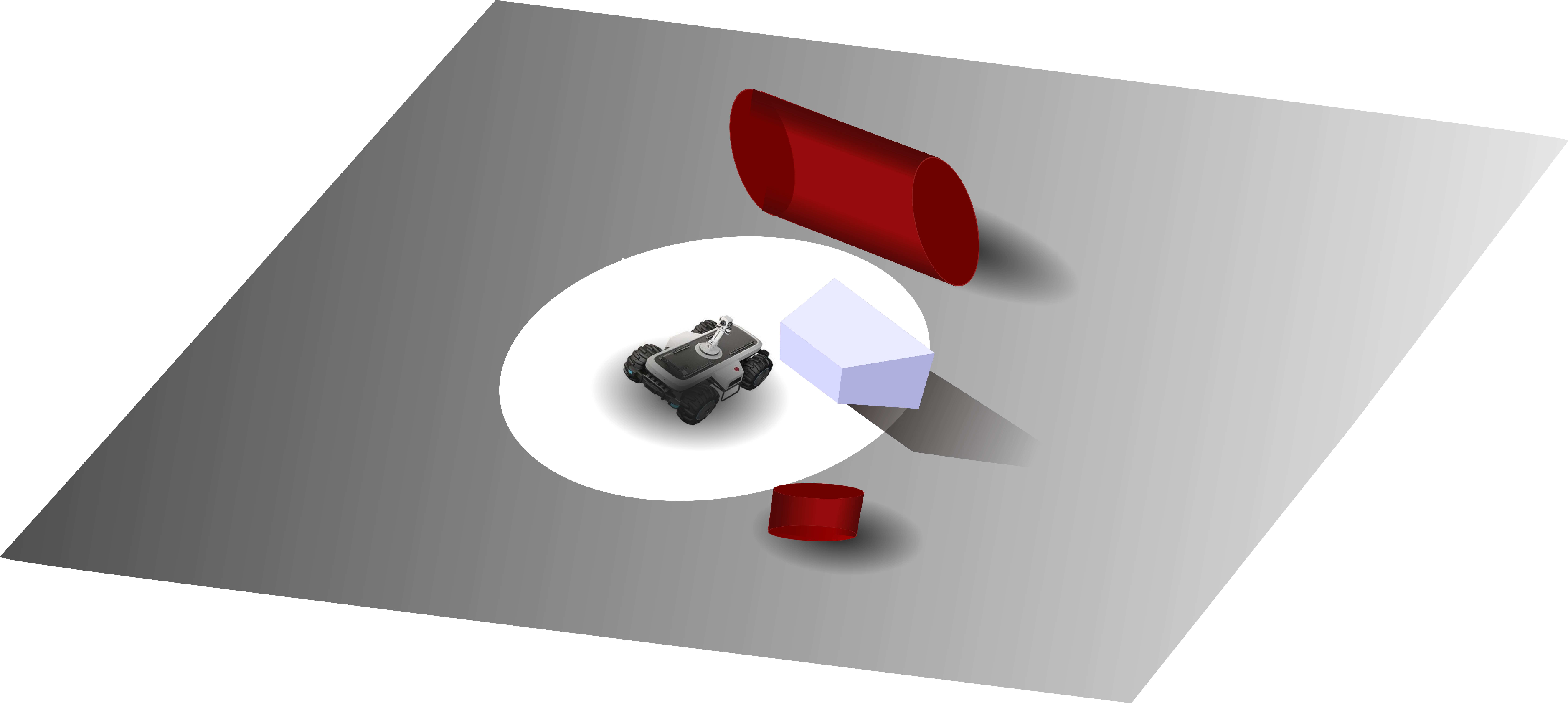}
    \caption{\footnotesize The white area around the robot shows its perception field. Any (parts of the) obstacles that fall within 
    this perception field are  known to the robot, while the robot is unaware of those obstacles that fall outside this field. 
    If no safety measures are considered, by positioning its center on the borders of the perception field, the body of the 
    robot may crash into the obstacles that are positioned close-by to the borders of its perception field and outside of it.}
    \label{fig:constraint_MPC}
\end{figure}

The objective function of the MPC problem for motion tracking per time step $\kappa$ 
is composed of two terms: The first term includes the offset of the state vector trajectory of the robot 
from the reference trajectory $\left\{\xref_{\kappa + 1}, \ldots, \xref_{\kappa + \Np} \right\}$ within the prediction window $\mathbb{P}_{\kappa} = \left\{\kappa + 1, \ldots, \kappa + \Np \right\}$ that is generated by the heuristic motion planning system. 
The second term of the objective function represents the kinetic energy of the robot 
and is included to impact the velocity vector of the robot, such that it improves the energy efficiency 
for the motion of the robot, and assists with the obstacle avoidance.
These terms are weighed using positive parameters $w_1<1$ and $w_2$, where the impact of the predictions that 
correspond to farther times in the future is discounted (due to being prone to larger 
estimation errors) by multiplying them by $w_1^k$. 
In other words, when $k$ is a larger time step within the prediction horizon, the 
weight of the corresponding term is smaller, because $w_1 < 1$. 
The TMPC problem for time step $\kappa$, with $\srob_{\kappa}$ and $\urob_{\kappa - 1}$ given, is  formulated 
within the prediction window $\mathbb{P}_{\kappa}$ via the following minimization problem:  
\begin{align}
    \label{eq:lin_obj} %
    \underset{\tildeu,\; \tildex}{\min} \;
    \bigg(\sum_{k = \kappa + 1}^{\kappa + \Np} w_1^{k/(\kappa + 1)} \norm{\srob_k - \xref_k}
    + w_2 \sum_{k = \kappa}^{\kappa + \Np - 1} \left({\urob_k}^\top\cdot \urob_k\right) \bigg)
\end{align}
subject to the following constraints, within the given prediction window $\mathbb{P}_{\kappa}$:
\begin{subequations}
\begin{align}
        & 
        \label{eq:lincon}
        \text{\eqref{eq:rob_dynamics} holds for updating the states of the robot}\\
        &
        \text{\eqref{eq:rob_dynamics} or a dynamic equation deduced from a Kalman filter holds for moving obstacles}\\
        &
        \smin \le \srob_{k} \le \smax \label{eq:lincon:state}\\
        &
        \umin \le \urob_{k - 1} \le \umax \label{eq:lincon:input}\\
        &
        \norm{\urob_{k - 1} - \urob_{k - 2}} \le \usm \label{eq:lincon:smooth}\\
        &
        \norm{\rrob_{k} - \rrob_{\kappa}} \le \rhoplan_{\kappa} - \rhosafe \label{eq:lincon:detected_zone}\\
        &
        \norm{\rrob_{k} - \rstobs(o_1)} \ge \rhosafe + \wrob_{k}, 
        \hspace{15 ex} \text{for} \; o_1 \in \isobs_{k}
        \label{eq:lincon:obs_static}\\
        &
        \norm{\rrob_{k} - \rdynobs_{k - 1}(o_2)} \ge \rhosafe + \wrob_{k} + \wdynobs_{k - 1}, 
        \hspace{7ex} \text{for} \; o_2 \in \idobs_{k}\; \text{and for $k > \kappa + 1$}
        \label{eq:lincon:obs_dynamic_1}
        \\
        &
        \norm{\rrob_{k} - \rdynobs_{k}(o_2)} \ge \rhosafe + \wrob_{k} + \wdynobs_{k}, 
        \hspace{7ex} \text{for} \; o_2 \in \idobs_{k}
        \label{eq:lincon:obs_dynamic_2}
        \\
        &
        \norm{\rrob_{k} - \rdynobs_{k + 1}(o_2)} \ge \rhosafe + \wrob_{k}+ \wdynobs_{k + 1}, 
        \hspace{7ex} \text{for} \; o_2 \in \idobs_{k}\; \text{and for $k < \kappa + \Np$}
        \label{eq:lincon:obs_dynamic} 
        \\
        &
        \urob_{k - 1} = \unomrob_{k - 1} + K_{k - 1} \left(\srob_{k - 1 } - \xref_{k - 1}\right)
        \label{eq:control_input_TMPC} 
\end{align}
\end{subequations}
where $\tildex = \left\{ \srob_{\kappa + 1}, \ldots, \srob_{\kappa + \Np} \right\}$ and $\tildeu = \left\{ \urob_{\kappa}, \ldots, \urob_{\kappa + \Nc - 1}, \ldots, \urob_{\kappa + \Np - 1}\right\}$, with $\urob_{\kappa + \Nc - 1} = \ldots = \urob_{\kappa + \Np - 1}$. 
Constraints \eqref{eq:lincon:state} and \eqref{eq:lincon:input} impose lower and upper 
limits on, respectively, the states and control inputs of the robot, where the symbol $\leq$ for vectors 
is executed element-wise on those vectors.  
Constraint \eqref{eq:lincon:smooth} limits the rate of the changes in the consecutive control inputs that are injected into the actuators of the robot, 
and guarantees smooth mechanical movements or dynamic variations for the actuators of the robot. 
Constraint \eqref{eq:lincon:detected_zone} ensures that the position of the center of the robot 
for the entire prediction window remains within a bounded zone with safe borders. 
This zone should embed the trajectory that is planned via the heuristic motion planning system of the robot as much as safety 
considerations allow for it. 
In other words, this constraint keeps the position of the center of the robot within a 
circle that is centered around the current position $\rrob_{\kappa}$ of the robot 
with a radius $\rhoplan_{\kappa} - \rhosafe$, where $\rhoplan_{\kappa}$ is the largest 
distance of $\rrob_{\kappa}$ from the planned path and $\rhosafe \geq \rhorob + \rhoobs$ 
is a parameter that is tuned based on how conservatively the safety guarantees are defined. 
Subtracting $\rhosafe$ from the planned radius ensures that when the center of the robot 
is positioned  on the borders of the safe zone, the robot will not crash into potential obstacles outside the zone. 
Figure~\ref{fig:constraint_MPC} illustrates the essence of including constraint \eqref{eq:lincon:detected_zone}. 
The variables $\isobs_{\kappa}$ and $\idobs_{\kappa}$ in \eqref{eq:lincon:obs_static}--\eqref{eq:lincon:obs_dynamic} are, respectively, the sets of all the static and dynamic obstacles that fall within the perception field of the robot at time step $k \in \mathbb{P}_{\kappa}$. 
Constraint \eqref{eq:lincon:obs_static} keeps the robot away from the detected static obstacles 
for all time steps $k \in \mathbb{P}_{\kappa}$ taking into account the impact of the 
disturbances on the position of the robot, by including the upper bound $w^{\text{rob}}_{k}$ for the disturbances. 
Constraints \eqref{eq:lincon:obs_dynamic_1}--\eqref{eq:lincon:obs_dynamic} 
prevent the robot from crashing into the dynamic obstacles that have been detected 
within the perception field of the robot at time step $\kappa$, by incorporating the impact of both the external disturbances that affect the position of the robot, with the upper bound $w^{\text{rob}}_{k}$ for $k \in \mathbb{P}_{\kappa}$, 
and the error in the perception of the estimated states of the dynamic obstacles, 
with the upper bounds $w^{\text{dyn,obs}}_{k - 1}$, $w^{\text{dyn,obs}}_{k}$, and $w^{\text{dyn,obs}}_{k + 1}$ 
for time steps $k - 1$, $k$, and $k+1$, respectively, where $k \in \mathbb{P}_{\kappa}\backslash \{ \kappa, \kappa + \Np\}$. 
Note that since a discrete-time problem is solved, per time step $k \in \mathbb{P}_{\kappa}\backslash \{ \kappa, \kappa + \Np\}$ 
the collision avoidance of the robot and the obstacle is enforced by providing the safe 
distance between their centers for the current time step and its immediate previous and next time steps, in order to reduce the risk of infeasibility. 
The reason for excluding time steps $\kappa$ and $\kappa + \Np$ from these constraints is the following: 
Since the robot does not hold any memories of the previous time steps (Assumption~\ref{ass:perception_field_memory}), 
at the beginning of the prediction horizon, i.e., at time step $k = \kappa$, it does not have access to the information of the dynamic obstacle for time step $k - 1 = \kappa - 1$. Therefore, it cannot estimate \eqref{eq:lincon:obs_dynamic_1}.  
Moreover, since the upper limit of the prediction horizon starting at time step 
$k = \kappa$ is time step $\kappa + \Np$, thus at time step $k = \kappa + \Np$ the robot does not 
have any information about the dynamic obstacle for time step $k + 1 = \kappa + \Np + 1$ and 
hence, cannot estimate \eqref{eq:lincon:obs_dynamic}. 
The upper bounds are time-varying to account for the errors accumulated within the prediction horizon, 
while preventing too much conservatism for TMPC \cite{BemporadMorari}.%
\smallskip

TMPC estimates or deploys nominal trajectories for the state and control inputs, and adjusts 
the control inputs online in order to ensure that, despite the external disturbances and perception errors, 
the realized state trajectory of the 
controlled system remains within a safe, feasible region, called the tube.
In our proposed architecture, the nominal state and control input trajectories for \eqref{eq:lin_obj}, subject to constraints \eqref{eq:lincon}--\eqref{eq:control_input_TMPC}, are instead those that have been 
injected into the motion tracking system via the motion planning system, i.e., $\xref_k$ and $\unomrob_k$. 
The control input adjusted online via TMPC is then determined via \eqref{eq:control_input_TMPC}, 
where the reference control input is adjusted using a control increment term 
that is determined based on a feedback from the system, i.e., the error between the realized and 
reference state trajectories. The gain $K_{k}$ used for $k + 1 \in \mathbb{P}_{\kappa}$ 
is determined per time step, as is common in TMPC literature, 
such that it stabilizes the error dynamics (which is usually done by linearizing 
\eqref{eq:rob_dynamics} per time step $k$ to obtain the dynamic and input matrices $A_k$ and $B_k$, 
respectively, and by enforcing the condition $f^{\text{SR}} \left( A_k + B_k K_k \right) < 1$ where  
$f^{\text{SR}}(\cdot)$ is the spectral radius function, i.e., a function that determines 
the largest eigenvalue of the input matrix, in this case $A_k + B_k K_k$). 
In particular, \eqref{eq:lincon:obs_dynamic_1}--\eqref{eq:lincon:obs_dynamic} ensure that the 
realized states remain within a safe tube that is dynamically tuned via the time-varying bounds 
$\wrob_{k}$ and $\wdynobs_{k}$, which we explain in the next section how to determine.%

In case the tracking TMPC problem is deemed infeasible, it 
calls back to the heuristic motion panning system to ask updating the reference trajectories. 
The optimization problem of TMPC is in general nonlinear and non-convex and may be solved by state-of-the-art global algorithms, e.g., genetic algorithm \cite{mitchell_1998} or pattern search \cite{torczon_1997}, using multiple starting points.

\subsubsection*{Dynamic adjustment of the tube in TMPC due to disturbances and perception errors} 

We decouple the evolution of the states of the robot due to its dynamics and due to the approaching dynamic obstacles, and independently incorporate the influence of these sources 
of uncertainties on the tube of TMPC.
The values of $\wrob_{k}$ and $\wdynobs_{k}$, computed at time step $\kappa$ and for $k \in \mathbb{P}_{\kappa}$, in the worst case are determined via the following  geometric sequences:
\begin{align}
    \label{eq:geomseq} 
    &\wrob_k = \Bar{w}^{\text{rob}} \sum_{i=0}^{k - \kappa - 1}(1-\xirob)^i \\
    \label{eq:geomseq2} 
    &\wdynobs_k = \Bar{w}^{\text{dyn,obs}} \sum_{i=0}^{k - \kappa - 1}(1-\xidyn)^i
\end{align}
where $\xirob \in [0, 1]$ and $\xidyn \in [0, 1]$ are the damping values for the deviation of the robot states and for the perception error regarding the position of the dynamic obstacles, respectively, and $\Bar{w}^{\text{rob}}$ and $\Bar{w}^{\text{dyn,obs}}$ are the upper bounds for the external disturbances that impact the robot states and for the perception error of the robot, respectively.
Based on these values, constraint tightening may be performed. Note that the damping vales should carefully be tuned to provide a balanced trade-off between increased robustness and reduced conservativeness for TMPC.%

\section*{Case study}
\label{ch:5cs}
Simulations were run to compare the performance of the proposed control architecture (called \gls{htmpc}, referring to integrated heuristic motion planning and TMPC) with two state-of-the-art methods,
\gls{hlrrt}\cite{chen2019horizon} and COLREGS \gls{apf}\cite{lyu2019colregs}. 
For all the simulations, the CPU used was a 4 core 2.5 GHz Intel® Core™ i7-4710MQ with 8GB of RAM memory and an integrated GPU of Intel® HD Graphics 4600.
The operating system was Ubuntu 18.04.6 LTS, a 64-bit OS, with open-source drivers, where applicable.
The simulations were done on MATLAB R2020b, where the parameters used have been made publicly available in the 4TU.ResearchData repository \cite{baglioni_dataset_2024}.%

\gls{hlrrt} is a path planning approach based on \gls{rrt}, 
which uses random sampling in its search space and builds up a tree 
with branches that connect the nearest points of the tree to each random sample, 
when this connection corresponds to a collision-free path. 
Once the target point is connected to the tree, the suitable path from the starting 
point to this target is selected.  
For enhanced computational efficiency, \gls{hlrrt} uses a horizon-based strategy 
to guide the exploration, where the sampling is biased toward the points that are closer to the horizon and/or to the target. Moreover, \gls{hlrrt} only checks the final candidate path for collision avoidance 
\cite{bruce2002real,yang2012robust}.%
 
COLREGS \gls{apf} refers to the deployment of \gls{apf} for navigation, complying with the rules of COLREGS \cite{lyu2019colregs}, i.e., international regulations for preventing collisions at sea. 
\gls{apf} steers the heading and velocity of the robot based on the vector that combines all attraction and repulsion (e.g., due to obstacles on the way) forces between the robot and its target.
The core aspect of COLREGS \gls{apf} used in this case study is a horizon-based collision-avoidance strategy, which makes the comparison with an MPC-based method more relevant.%
 
A square-shaped environment of size $14~\text{m} \times 14~\text{m}$ was simulated, 
considering \textbf{case~1} with $10$ scenarios, each including $6$ static and $5$ dynamic obstacles, 
and \textbf{case~2} with $10$ other scenarios, each including $8$ static and $8$ dynamic obstacles. 
In \textbf{case~2} in particular initial configurations and kinematics were designed for the obstacles such that a temporarily infeasible problem would appear for the robot.
The scenarios were carefully designed to ensure that reaching the destination for the robot was always feasible in the long term.
The random variables (e.g., the external disturbance affecting the position of the robot) were different 
among the scenarios for each case (for details see the data repository \cite{baglioni_dataset_2024}).
Note that while the error in perceiving the position of dynamic obstacles and the deviation of the states 
of the robot due to external disturbances were bounded and randomly generated, the same values were used for different control methods in order to make the comparisons fair.%

The simulations were run following two setups: 
(1) A computation budget ($0.15$~s per decision making for MPC and \gls{hlrrt}, while \gls{apf} does not in practice need this time budget, as it solves the problem almost in real time) was considered, where the simulations were terminated as soon as the budget was exhausted. 
(2) The three approaches ran until either the target was reached by the robot or reaching the 
target was deemed infeasible. 
The comparisons of the performance among the three approaches were with regards to the rate of success of the three methods (i.e., 
whether or not the robot reaches the target without any collisions and without falling into a livelock, e.g., circling) and the length of the path taken by the robot to the target. 
In setup (2), the overall mission time (i.e., the time taken by the robot to reach its target) was  also compared.%

To simulate the motion of each dynamic obstacle $o$,
the following nonlinear equations were considered: 
\begin{align}
    \label{eq:obs:dynamics_x,y}
    &\xobs{dyn}_{\kappa + 1} (o) = \xobs{dyn}_{\kappa}(o) + \text{RK} \left(\vxobs{\kappa}(o),\cctrl \right),
    &\vxobs{\kappa + 1}(o) = \vxobs{\kappa}(o) + \text{RK}\left( \alpha (\xat(o)- \xobs{dyn}_{\kappa}(o)),\cctrl \right), \\
    &\yobs{dyn}_{\kappa + 1}(o) = \yobs{dyn}_{\kappa}(o) + \text{RK}\left(\vyobs{\kappa}(o),\cctrl \right), 
    &\vyobs{\kappa + 1}(o) = \vyobs{\kappa}(o) + \text{RK}\left(\beta (\yat(o)-\yobs{dyn}_{\kappa}(o)),\cctrl \right)  
\end{align} 
with $\text{RK}(\cdot, \cctrl)$ Runge-Kutta 3/8 operator that integrates the given variable across one sampling time $\cctrl$. 
The initial positions and the velocities of the obstacles were sampled based on a uniform distribution, 
and the motions were around fixed attraction points with coordinates $[\xat(o),\yat(o)]^{\top}$ 
per obstacle $o$. 
For the constant multipliers $\alpha$ and $\beta$ we considered: 
\begin{align*}
    \alpha = 0.2\frac{1 + 4\eta}{v^{\text{rob,max}}_x - v^{\text{rob,min}}_x + \norm{\xobs{dyn}_{0} (o) - \xat(o)}}, 
    \quad
    \beta = 0.2\frac{1 + 4\eta}{v^{\text{rob,max}}_y - v^{\text{rob,min}}_y + \norm{\yobs{dyn}_{0}(o) - \yat(o)}}   
\end{align*}
with $\eta \in [0 , 1]$ a random number per obstacle that is sampled from a uniform distribution, and $v^{\text{rob,max}}_x$, $v^{\text{rob,min}}_x$, $v^{\text{rob,max}}_y$, $v^{\text{rob,min}}_y$ 
the maximum and minimum velocities of the robot in the $x$ and $y$ directions, respectively.
These choices allow for relative velocities for the robot and the dynamic obstacles that result in obstruction of the path of the robot, thus evaluating the given approaches based on relevant, meaningful simulations. 
We assume that the robot uses a filter to estimate the motion of the dynamic obstacles, simply using linear approximation for the velocities.
Such an approximation, 
considering the prediction horizon of $\Np = 5$ used in the case study, 
results in a bounded error of maximum $3.57 \%$, which is acceptable for the simulations.%

\section*{Simulation results}
\label{ch:6sr}
In this section the results of the simulations are presented. 
Due to the large number of experiments, we have presented only a few representative cases:   
Figures~\ref{fig:MPC_simple}-\ref{fig:MPC_complex}, 
Figures~\ref{fig:RRT_simple}-\ref{fig:RRT_complex},  
and Figures~\ref{fig:APF_simple}-\ref{fig:APF_complex} 
correspond to deploying, respectively, \gls{htmpc}, \gls{hlrrt}, and \gls{apf} for \textbf{case~1} and \textbf{case~2}. 
The dash-dotted blue and solid green trajectories in the plots correspond to, respectively, setup 1 and setup 2.
Static and dynamic obstacles are illustrated as solid black and red circles, respectively.
In the plots on the right-hand side of these figures, 
the realized distance of the robot from the closest obstacle during the simulation, 
as well as the minimum safety radius (black dashed lines) are shown.%

Tables~\ref{tab:sn_results} and \ref{tab:cn_results} show the results 
for the path length and mission time of the robot for \textbf{case~1} and \textbf{case~2}, respectively. 
A dash symbol is used to indicate failure, i.e., the robot did not reach the target, due to either collisions or falling into livelocks.

\begin{table}[!htb]
    \centering
    \caption{\footnotesize Results in terms of the path length (m) and the mission time (s) 
    for \textbf{case~1}, setup 1 (left-hand side table) and setup 2 (right-hand side table).}
    \label{tab:sn_results}
    \begin{tabular}{c cc cc cc} \toprule
    & \multicolumn{2}{c}{\footnotesize{\gls{htmpc}}} & \multicolumn{2}{c}{\footnotesize{\gls{hlrrt}}} & \multicolumn{2}{c}{\footnotesize{\gls{apf}}} \\ \cmidrule(r){2-3} \cmidrule(r){4-5} \cmidrule(r){6-7}
        \footnotesize{scenario}    & \footnotesize{path} & \footnotesize{time}  & \footnotesize{path} & \footnotesize{time} & \footnotesize{path} & \footnotesize{time} \\ \midrule
      \footnotesize{1} & \footnotesize{18.8} & \footnotesize{49.8} & \footnotesize{-}    & \footnotesize{-}    & \footnotesize{-}   & \footnotesize{-}  \\
      \footnotesize{2} & \footnotesize{20.1} & \footnotesize{55.7} & \footnotesize{20.0} & \footnotesize{44.5} & \footnotesize{-}   & \footnotesize{-} \\
      \footnotesize{3} & \footnotesize{22.3} & \footnotesize{50.5} & \footnotesize{-}   & \footnotesize{-}    & \footnotesize{-}   & \footnotesize{-}  \\
      \footnotesize{4} & \footnotesize{17.9} & \footnotesize{53.3} & \footnotesize{19.3} & \footnotesize{40.2} & \footnotesize{-}   & \footnotesize{-}  \\
      \footnotesize{5} & \footnotesize{16.5} & \footnotesize{41.1} & \footnotesize{19.8} & \footnotesize{44.2} & \footnotesize{-}   & \footnotesize{-}  \\
      \footnotesize{6} & \footnotesize{21.0} & \footnotesize{55.2} & \footnotesize{21.9} & \footnotesize{52.1} & \footnotesize{-}   & \footnotesize{-}  \\
      \footnotesize{7} & \footnotesize{15.2} & \footnotesize{32.3} & \footnotesize{15.0} & \footnotesize{30.9} & \footnotesize{15.4} & \footnotesize{30.8}  \\
      \footnotesize{8} & \footnotesize{22.4} & \footnotesize{49.7} & \footnotesize{-}    & \footnotesize{-}    & \footnotesize{-}   & \footnotesize{-}  \\
      \footnotesize{9} & \footnotesize{20.0} & \footnotesize{59.5} & \footnotesize{24.9} & \footnotesize{22.9} & \footnotesize{-}   & \footnotesize{-}  \\
      \footnotesize{10} & \footnotesize{17.8} & \footnotesize{39.3} & \footnotesize{19.2} & \footnotesize{39.2} & \footnotesize{-}   & \footnotesize{-}  \\ \midrule
      \footnotesize{mean}  & \footnotesize{19.2} & \footnotesize{48.6} & \footnotesize{20.0} & \footnotesize{44.2} & \footnotesize{15.4} & \footnotesize{30.8} \\
      \footnotesize{standard deviation} & \footnotesize{2.40} & \footnotesize{8.50} & \footnotesize{2.99} & \footnotesize{8.94} & \footnotesize{-}   & \footnotesize{-}  \\ \bottomrule
    \end{tabular}
    \quad
    \begin{tabular}{c cc cc cc} \toprule
    & \multicolumn{2}{c}{\footnotesize{\gls{htmpc}}} & \multicolumn{2}{c}{\footnotesize{\gls{hlrrt}}} & \multicolumn{2}{c}{\footnotesize{\gls{apf}}} \\ \cmidrule(r){2-3} \cmidrule(r){4-5} \cmidrule(r){6-7}
        \footnotesize{scenario}    & \footnotesize{path} & \footnotesize{time}  & \footnotesize{path} & \footnotesize{time} & \footnotesize{path} & \footnotesize{time} \\ \midrule
       \footnotesize{1} & \footnotesize{15.5} & \footnotesize{40.6} & \footnotesize{26.4} & \footnotesize{50.7} & \footnotesize{-}   & \footnotesize{-}  \\
      \footnotesize{2} & \footnotesize{16.8} & \footnotesize{37.1} & \footnotesize{19.4} & \footnotesize{41.2} & \footnotesize{-}   & \footnotesize{-} \\
      \footnotesize{3} & \footnotesize{21.0} & \footnotesize{52.1} & \footnotesize{24.0} & \footnotesize{46.3} & \footnotesize{-}   & \footnotesize{-}  \\
      \footnotesize{4} & \footnotesize{16.8} & \footnotesize{37.6} & \footnotesize{18.7} & \footnotesize{35.0} & \footnotesize{-}   & \footnotesize{-}  \\
      \footnotesize{5} & \footnotesize{16.5} & \footnotesize{42.7} & \footnotesize{17.6} & \footnotesize{32.5} & \footnotesize{-}   & \footnotesize{-}  \\
       \footnotesize{6} & \footnotesize{18.2} & \footnotesize{38.1} & \footnotesize{21.0} & \footnotesize{44.5} & \footnotesize{-}   & \footnotesize{-}  \\
       \footnotesize{7} & \footnotesize{15.0} & \footnotesize{31.8} & \footnotesize{16.0} & \footnotesize{30.9} & \footnotesize{15.4} & \footnotesize{30.8}  \\
       \footnotesize{8} & \footnotesize{19.1} & \footnotesize{50.6} & \footnotesize{22.9} & \footnotesize{50.4} & \footnotesize{-}   & \footnotesize{-}  \\
       \footnotesize{9} & \footnotesize{18.2} & \footnotesize{41.1} & \footnotesize{20.5} & \footnotesize{42.5} & \footnotesize{-}   & \footnotesize{-}  \\
       \footnotesize{10} & \footnotesize{16.9} & \footnotesize{41.9} & \footnotesize{18.3} & \footnotesize{33.3} & \footnotesize{-}   & \footnotesize{-}  \\ \midrule
      \footnotesize{mean}  & \footnotesize{17.4} & \footnotesize{41.4} & \footnotesize{20.5} & \footnotesize{40.7} & \footnotesize{15.4} & \footnotesize{30.8} \\
      \footnotesize{standard deviation} & \footnotesize{1.78} & \footnotesize{6.13} & \footnotesize{3.18} & \footnotesize{7.74} & \footnotesize{-}   & \footnotesize{-}  \\ \bottomrule
    \end{tabular}
\end{table}

\begin{table}[!htb]
    \centering
    \caption{\footnotesize Results in terms of the path length (m) and the mission time (s) 
    for \textbf{case~2}, setup 1 (left-hand side table) and setup 2 (right-hand side table)..}
    \label{tab:cn_results}
    \begin{tabular}{c cc cc cc} \toprule
         & \multicolumn{2}{c}{\footnotesize{\gls{htmpc}}} & \multicolumn{2}{c}{\footnotesize{HL-RRT*}} & \multicolumn{2}{c}{\footnotesize{APF}} \\ \cmidrule(r){2-3} \cmidrule(r){4-5} \cmidrule(r){6-7}
        \footnotesize{scenario}    & \footnotesize{path} & \footnotesize{time}  & \footnotesize{path} & \footnotesize{time} & \footnotesize{path} & \footnotesize{time} \\ \midrule
      \footnotesize{1} & \footnotesize{-}   & \footnotesize{-}    & \footnotesize{-}   & \footnotesize{-}   & \footnotesize{-}   & \footnotesize{-}  \\
      \footnotesize{2} & \footnotesize{-}   & \footnotesize{-}    & \footnotesize{-}   & \footnotesize{-}   & \footnotesize{-}   & \footnotesize{-}  \\
      \footnotesize{3} & \footnotesize{19.1} & \footnotesize{47.8}  & \footnotesize{-}   & \footnotesize{-}    & \footnotesize{-}   & \footnotesize{-}  \\
      \footnotesize{4} & \footnotesize{-}   & \footnotesize{-}    & \footnotesize{-}   & \footnotesize{-}   & \footnotesize{-}   & \footnotesize{-}  \\
      \footnotesize{5} & \footnotesize{-}   & \footnotesize{-}    & \footnotesize{-}   & \footnotesize{-}   & \footnotesize{-}   & \footnotesize{-}  \\
      \footnotesize{6} & \footnotesize{20.9} & \footnotesize{51.0} & \footnotesize{-}   & \footnotesize{-}    & \footnotesize{-}   & \footnotesize{-}  \\
      \footnotesize{7} & \footnotesize{-}   & \footnotesize{-}    & \footnotesize{-}   & \footnotesize{-}   & \footnotesize{-}   & \footnotesize{-}  \\
      \footnotesize{8} & \footnotesize{-}   & \footnotesize{-}    & \footnotesize{-}   & \footnotesize{-}   & \footnotesize{-}   & \footnotesize{-}  \\
      \footnotesize{9} & \footnotesize{-}   & \footnotesize{-}    & \footnotesize{-}   & \footnotesize{-}   & \footnotesize{-}   & \footnotesize{-}  \\
      \footnotesize{10} & \footnotesize{-}   & \footnotesize{-}    & \footnotesize{-}   & \footnotesize{-}   & \footnotesize{-}   & \footnotesize{-}  \\ \midrule
      \footnotesize{mean}  & \footnotesize{20.0} & \footnotesize{49.4} & \footnotesize{-}   & \footnotesize{-}    & \footnotesize{-}   & \footnotesize{-}  \\
      \footnotesize{standard deviation} & \footnotesize{1.26} & \footnotesize{2.26} & \footnotesize{-}   & \footnotesize{-}    & \footnotesize{-}   & \footnotesize{-}  \\ \bottomrule
    \end{tabular}
    \quad
    \begin{tabular}{c cc cc cc} \toprule
      & \multicolumn{2}{c}{\footnotesize{\gls{htmpc}}} & \multicolumn{2}{c}{\footnotesize{HL-RRT*}} & \multicolumn{2}{c}{\footnotesize{APF}} \\ \cmidrule(r){2-3} \cmidrule(r){4-5} \cmidrule(r){6-7}
        \footnotesize{scenario}    & \footnotesize{path} & \footnotesize{time}  & \footnotesize{path} & \footnotesize{time} & \footnotesize{path} & \footnotesize{time} \\ \midrule
      \footnotesize{1} & \footnotesize{19.2} & \footnotesize{50.7}  & \footnotesize{-}   & \footnotesize{-}    & \footnotesize{-}   & \footnotesize{-}  \\
      \footnotesize{2} & \footnotesize{16.2} & \footnotesize{43.1}  & \footnotesize{-}   & \footnotesize{-}    & \footnotesize{-}   & \footnotesize{-}  \\
      \footnotesize{3} & \footnotesize{17.4} & \footnotesize{41.0}  & \footnotesize{-}   & \footnotesize{-}    & \footnotesize{-}   & \footnotesize{-} \\
      \footnotesize{4} & \footnotesize{18.4} & \footnotesize{45.0}  & \footnotesize{-}   & \footnotesize{-}    & \footnotesize{-}   & \footnotesize{-} \\
      \footnotesize{5} & \footnotesize{19.5} & \footnotesize{52.0}  & \footnotesize{-}   & \footnotesize{-}    & \footnotesize{-}   & \footnotesize{-} \\
      \footnotesize{6} & \footnotesize{17.2} & \footnotesize{48.7}  & \footnotesize{-}   & \footnotesize{-}    & \footnotesize{-}   & \footnotesize{-} \\
      \footnotesize{7} & \footnotesize{18.4} & \footnotesize{46.8}  & \footnotesize{-}   & \footnotesize{-}    & \footnotesize{-}   & \footnotesize{-} \\
      \footnotesize{8} & \footnotesize{-}   & \footnotesize{-}    & \footnotesize{-}   & \footnotesize{-}   & \footnotesize{-}   & \footnotesize{-}  \\
      \footnotesize{9} & \footnotesize{-}   & \footnotesize{-}    & \footnotesize{-}   & \footnotesize{-}   & \footnotesize{-}   & \footnotesize{-}  \\
      \footnotesize{10} & \footnotesize{-}   & \footnotesize{-}    & \footnotesize{-}   & \footnotesize{-}   & \footnotesize{-}   & \footnotesize{-}  \\ \midrule
      \footnotesize{mean}  & \footnotesize{18.0} & \footnotesize{46.8}  & \footnotesize{-}   & \footnotesize{-}    & \footnotesize{-}   & \footnotesize{-} \\
      \footnotesize{standard deviation} & \footnotesize{1.16} & \footnotesize{4.01}  & \footnotesize{-}   & \footnotesize{-}    & \footnotesize{-}   & \footnotesize{-} \\ \bottomrule
    \end{tabular}
\end{table}


\section*{Discussion of the results}
\label{ch:7d}

\begin{figure}
    \begin{subfigure}{0.3\textwidth}
        \centering        \includegraphics[width=\linewidth]{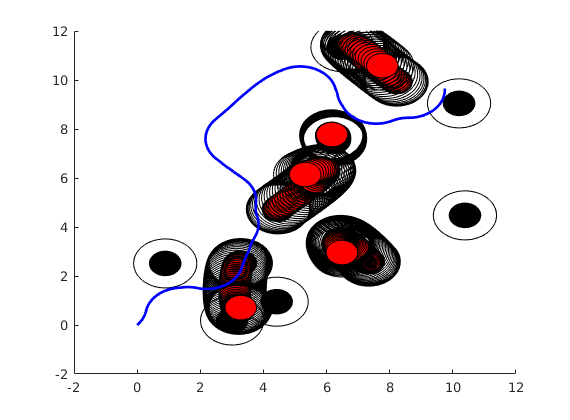}
        \caption{\footnotesize Trajectory towards the target (shown in blue) generated by \gls{apf} 
        following sub-optimal re-tuning of its hyper-parameters.}
        \label{fig:apf_modified}
    \end{subfigure}
    \hspace{1ex}
    \begin{subfigure}{0.3\textwidth}
        \centering
        \includegraphics[width=\linewidth]{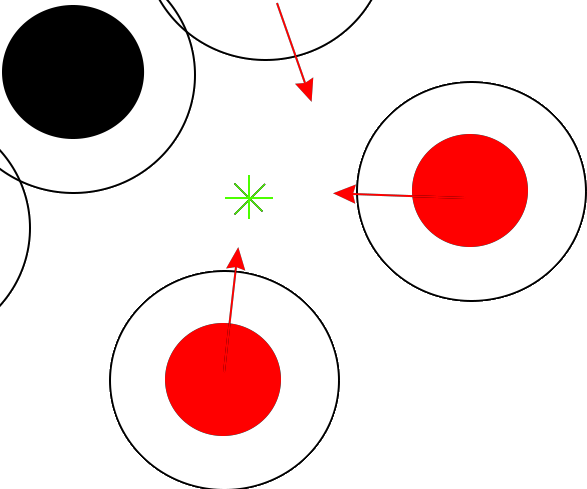}
        \caption{\footnotesize Crushing motion scenario.
        Three dynamic obstacles (with arrows showing their movement) move towards the robot (star).}
        \label{fig:crushing}
    \end{subfigure}
    \hspace{1ex}
    \begin{subfigure}{0.35\textwidth}
        \centering
        \includegraphics[width=\linewidth]{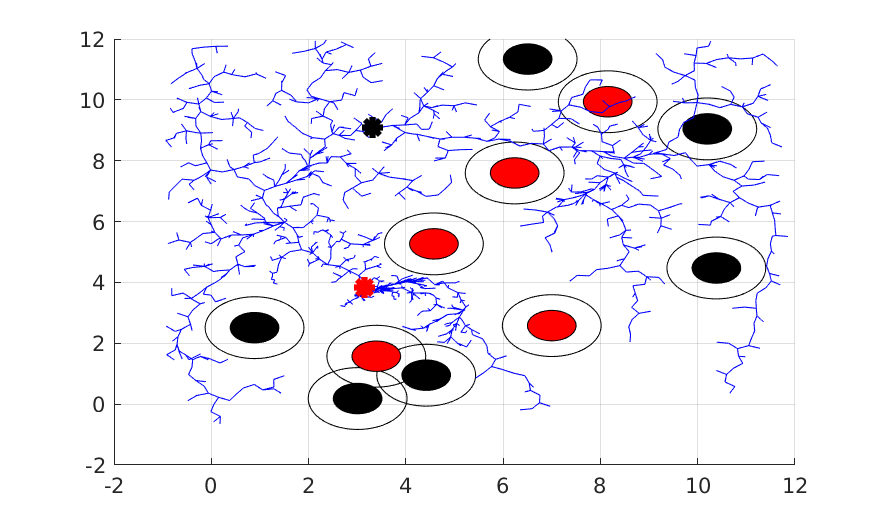}
        \caption{\footnotesize Tree structure (blue) of the \gls{hlrrt} with a high computational budget, at the time of the berth in Figure~\ref{fig:sn_rrt_path}.
        Collision checking is done from the current position (red star) to the end of the horizon (black star).}
        \label{fig:snn_rrt_berth}
    \end{subfigure}
    \caption{\footnotesize Particular conditions in the simulations of the case study.}
\end{figure}

\begin{sidewaysfigure*}
    \centering
    \begin{tabular}{@{}c@{}}
        \subfloat[]{\includegraphics[width=.41\textwidth]{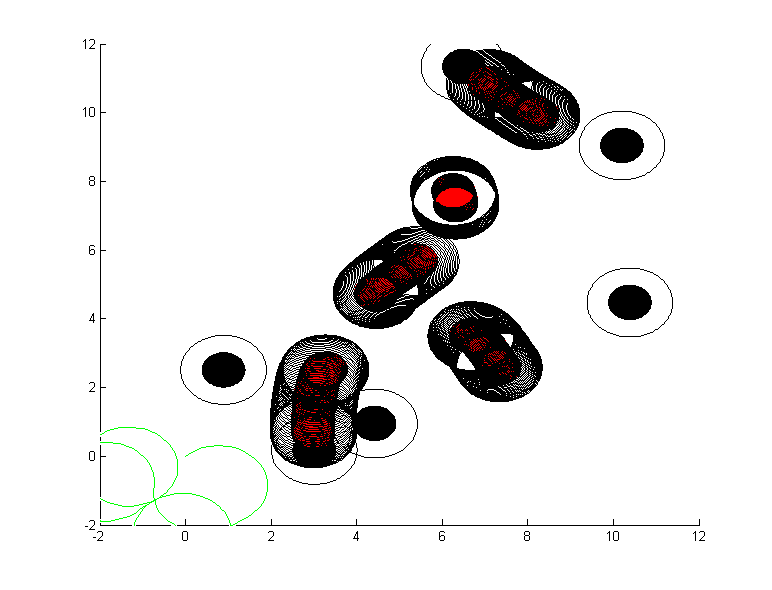}
        \label{fig:sn_apf_path}}
        \hfil
    \end{tabular}
    \begin{tabular}{@{}c@{}}
        \centering
        \subfloat[]{\includegraphics[width=.55\textwidth]{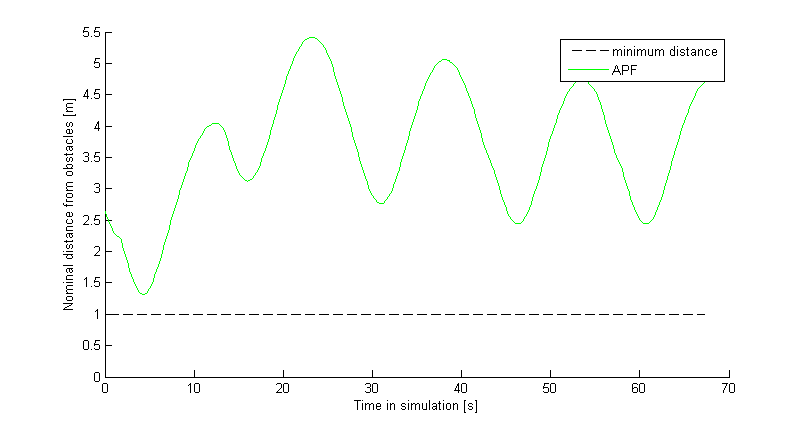}
        \label{fig:sn_apf_dist}}
    \end{tabular}
    \caption{\footnotesize (a) Sample robot paths for \textbf{case~1} for \textbf{\gls{apf}} and (b) the distance of the robot from the closest obstacles.}
    \label{fig:APF_simple}

\hspace{2ex}

    \centering
    \begin{tabular}{@{}c@{}}
        \centering
        \subfloat[]{\includegraphics[width=.41\textwidth]{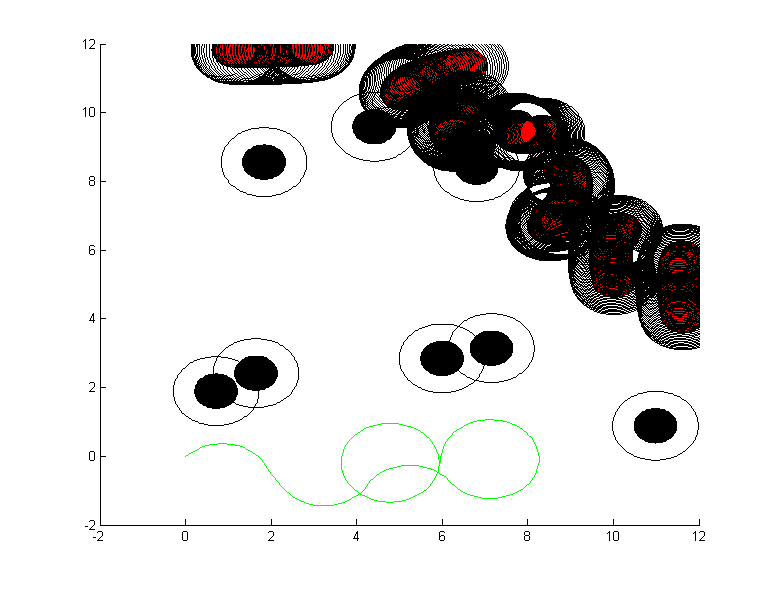}
        \label{fig:cn_apf_path}}
        \hfil
    \end{tabular}
    \begin{tabular}{@{}c@{}}
        \centering
        \subfloat[]{\includegraphics[width=.55\textwidth]{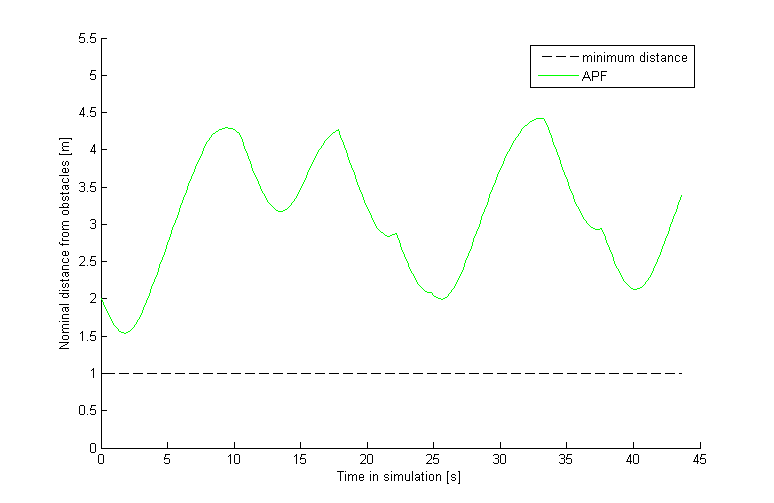}
        \label{fig:cn_apf_dist}}
    \end{tabular}
    \caption{\footnotesize a) Sample robot paths for \textbf{case~2} for \textbf{\gls{apf}} and (b) the distance of the robot from the closest obstacles.}
    \label{fig:APF_complex}
\end{sidewaysfigure*}

In this section, the results are discussed. 
In general, \gls{apf} failed to perform satisfactorily in both \textbf{case~1} and \textbf{case~2}, showing only a single success from the 10 scenarios for both setup 1 and setup 2.
Failures occurred because the robot got stuck in an $8$-shaped path (see Figures~\ref{fig:sn_apf_path} and~\ref{fig:cn_apf_path}).
This turned out to be related to the tuning of \gls{apf}, 
since after re-tuning it, the livelock disappeared and a trajectory towards the target was found (see Figure~\ref{fig:apf_modified}).
This, however, shows the very high sensitivity of \gls{apf} to its hyper-parameters, 
making it a less suitable approach for robots in SaR or other dynamic environments with limited information.%

From Table~\ref{tab:sn_results}, \gls{htmpc} had a higher success rate than \gls{hlrrt} for setup 1.
All failures of \gls{hlrrt} were due to crashing of the robot into obstacles, especially in a specific scenario designed to increase the risk of crashing (see Figure~\ref{fig:crushing}),
where the robot was placed between two dynamic obstacles or one dynamic and one static obstacle. 
In this case, \gls{hlrrt}  did not find an alternative lower cost path in time that keeps the robot safe.
Instead, \gls{htmpc} avoided crashing into the obstacles in this challenging scenario,
by either considering an obstacle belt (as a result of modeling the motion of dynamic obstacles within the prediction horizon) to avoid, or setting a reference trajectory outside the unsafe region. 
For setup 2, however, \gls{hlrrt} always handled this high-risk scenario,
but with a high computational cost. 
In fact, using \gls{hlrrt} the robot only moves across a path that leads to a lower cost, in this case a path that is closer to the target.
Thus, unless a node was found closer to the target than the current position of the robot, it simply refused to move.
Since in setup 2 there was always time to find such a node, \gls{hlrrt} showed a $100\%$ success rate.%

\begin{sidewaysfigure*}
    \centering
    \begin{tabular}{@{}c@{}}
        \subfloat[]{\includegraphics[width=.41\textwidth]{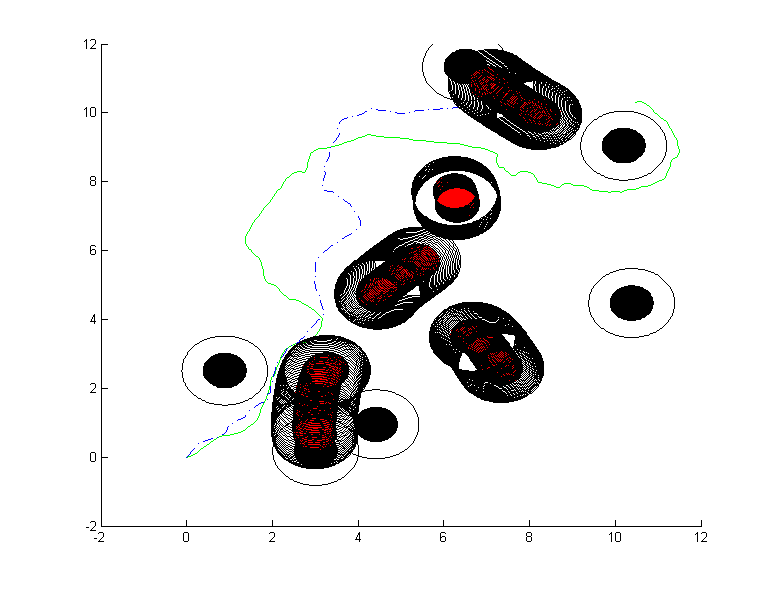}
        \label{fig:sn_rrt_path}}
        \hfil
    \end{tabular}
    \begin{tabular}{@{}c@{}}
        \centering
        \subfloat[]{\includegraphics[width=.55\textwidth]{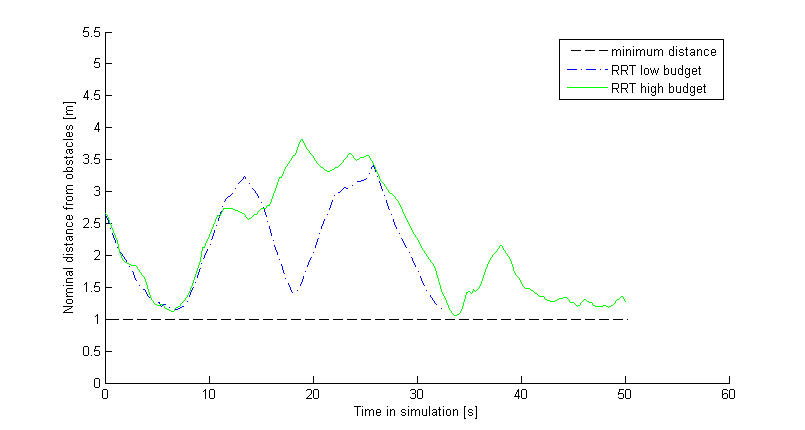}
        \label{fig:sn_rrt_dist}}
    \end{tabular}
    \caption{\footnotesize (a) Sample robot paths for \textbf{case~1}, setups 1 and 2 (called low and high budget respectively) for \textbf{\gls{hlrrt}} and (b) the distance of the robot from the closest obstacles.}
    \label{fig:RRT_simple}

\hspace{2ex}

    \centering
    \begin{tabular}{@{}c@{}}
        \centering
        \subfloat[]{\includegraphics[width=.41\textwidth]{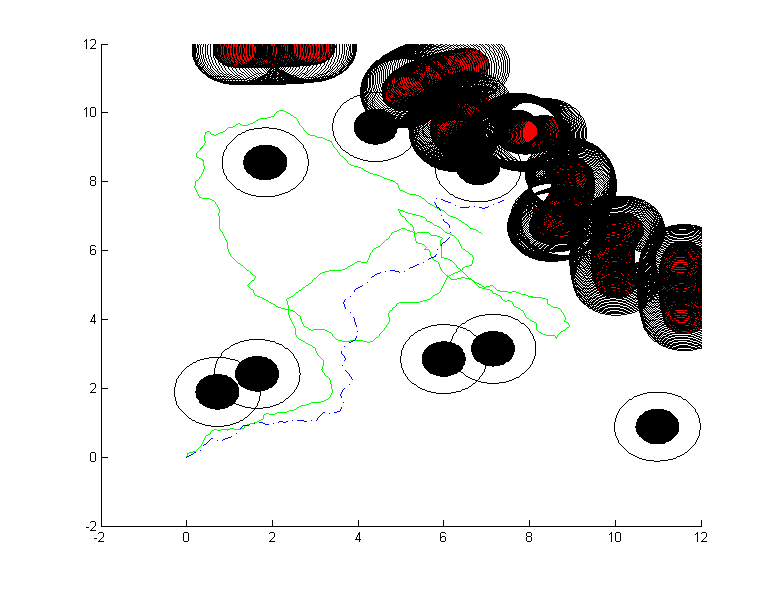}
        \label{fig:cn_rrt_path}}
        \hfil
    \end{tabular}
    \begin{tabular}{@{}c@{}}
        \centering
        \subfloat[]{\includegraphics[width=.55\textwidth]{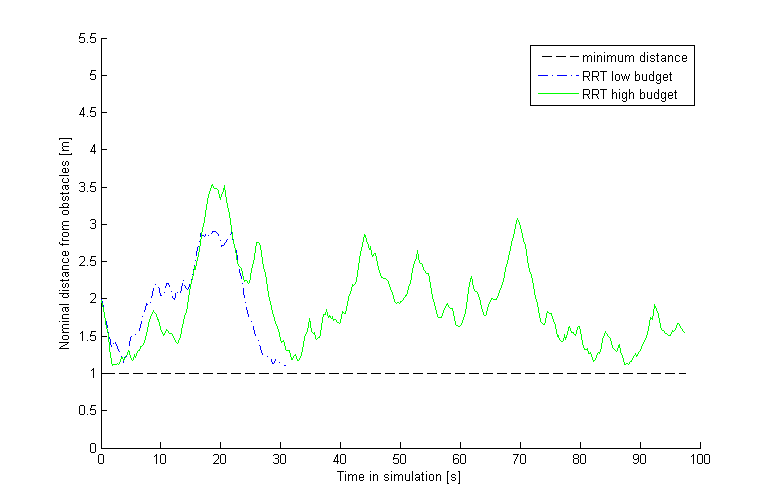}
        \label{fig:cn_rrt_dist}}
    \end{tabular}
    \caption{\footnotesize (a) Sample robot paths for \textbf{case~2}, setups 1 and 2 (called low and high budget respectively) for \textbf{\gls{hlrrt}} and (b) the distance of the robot from the closest obstacles.}
    \label{fig:RRT_complex}
\end{sidewaysfigure*}

Comparing the path lengths, \gls{hlrrt} generally took a longer path than \gls{htmpc}, while the path of \gls{hlrrt} was generally shorter in setup 2 compared to the same approach used in setup 1.
In a few cases, \gls{hlrrt} took a slightly shorter path than \gls{htmpc}, due to its random nature.
See the path made by \gls{hlrrt} in Figure~\ref{fig:sn_rrt_path}, where a wide berth 
is made to avoid the middle obstacles, partly due to the sampling and partly because of the local cost propagation in the tree, a limitation of \gls{rrt}.
This is confirmed considering the tree that was used at the time of the berth (see Figure~\ref{fig:snn_rrt_berth}).%

\begin{sidewaysfigure*}
    \centering
    \begin{tabular}{@{}c@{}}
        \subfloat[]{\includegraphics[width=.41\textwidth]{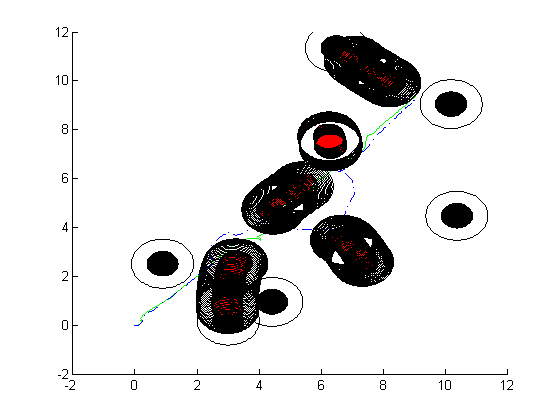}
        \label{fig:sn_mpc_path}}
        \hfil
    \end{tabular}
    \begin{tabular}{@{}c@{}}
        \centering
        \subfloat[]{\includegraphics[width=.55\textwidth]{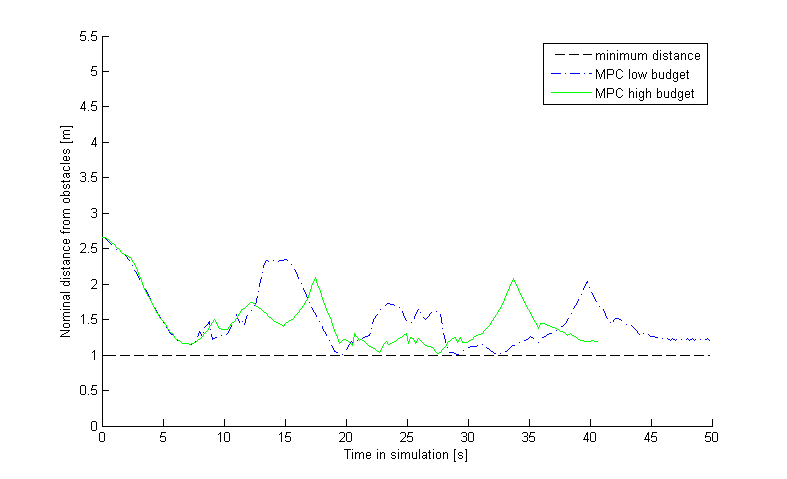}
        \label{fig:sn_mpc_dist}}
    \end{tabular}
    \caption{\footnotesize (a) Sample robot paths for \textbf{case~1}, setups 1 and 2 (called low and high budget respectively) for the \textbf{proposed control architecture} and (b) the distance of the robot from the closest obstacles.}
    \label{fig:MPC_simple}

\hspace{2ex}

    \centering
    \begin{tabular}{@{}c@{}}
        \centering
        \subfloat[]{\includegraphics[width=.41\textwidth]{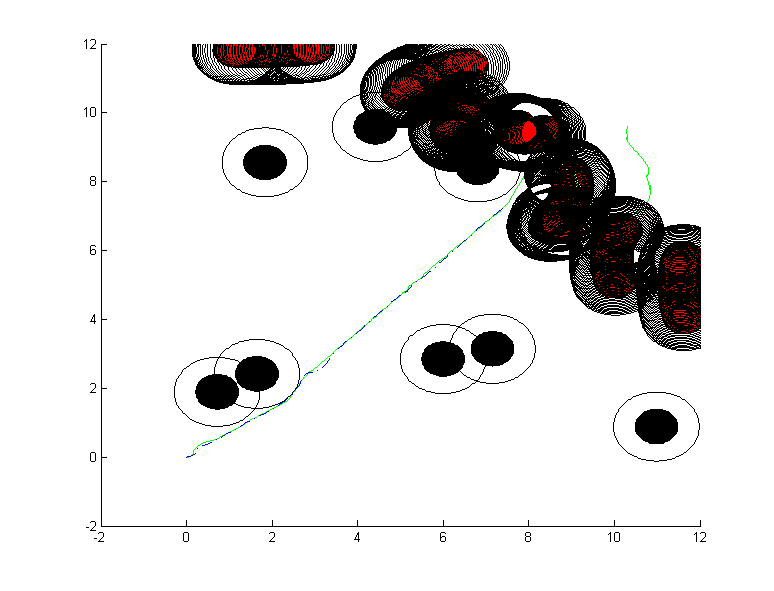}
        \label{fig:cn_mpc_path}}
        \hfil
    \end{tabular}
    \begin{tabular}{@{}c@{}}
        \centering
        \subfloat[]{\includegraphics[width=.55\textwidth]{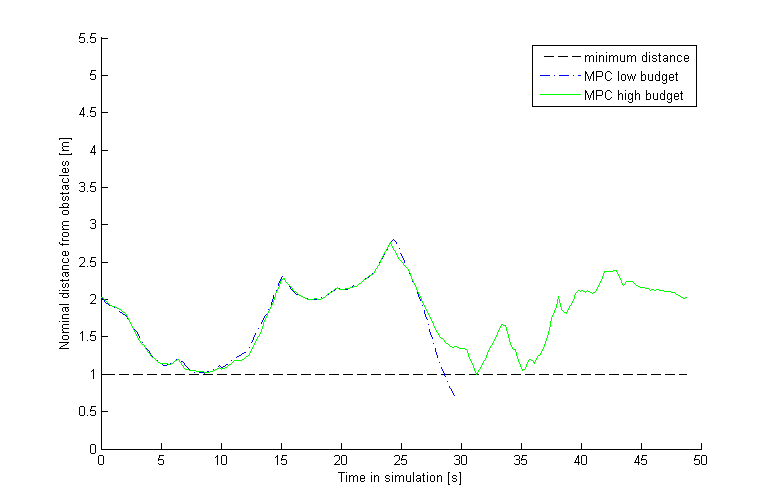}
        \label{fig:cn_mpc_dist}}
    \end{tabular}
    \caption{\footnotesize (a) Sample robot paths for \textbf{case~2}, setups 1 and 2 (called low and high budget respectively) for the \textbf{proposed control architecture} and (b) the distance of the robot from the closest obstacles.}
    \label{fig:MPC_complex}
\end{sidewaysfigure*}

For \gls{htmpc}, setup 1 resulted in paths that were approximately $10\%$ longer on average than for the same approach used in setup 2.
This can be explained via Figure~\ref{fig:sn_mpc_path}, where a small berth for setup 1 is observed that is avoided in setup 2 (see Figure~\ref{fig:cn_mpc_path}).
This is linked to the limited knowledge about dynamic obstacles within a limited computational window.
In fact, for setup 2 \gls{htmpc} found a lower cost path by slightly changing the course and speeding up, and it almost always found a local minimum, 
while the optimization in setup 1 sometimes stopped prematurely.%

For the mission time, in setup~1 in various cases when \gls{hlrrt} has found a path, this path has generally resulted in a smaller mission time than with \gls{htmpc}. 
In setup 2, in almost half of the cases \gls{hlrrt} wins in achieving a smaller mission time, while in the other cases \gls{htmpc} wins, with, on average, $11\%$ reduced time for the winning approach in both cases.
Comparing, for instance, Figures~\ref{fig:MPC_simple} and \ref{fig:RRT_simple}, it is clear that 
\gls{htmpc} has opted for the shortest path to the target, but since this requires 
moving closely to various obstacles, it may have compromised its speed for remaining crash-free, 
whereas by taking a longer path, \gls{hlrrt} has avoided the obstacles significantly. 
This is mainly due to the formulation of the optimization problem for 
\gls{htmpc} (see \eqref{eq:lin_obj}), where no explicit term for reducing the mission time has been considered in 
the objective function, but rather the controller is asked to minimize its distance from 
a reference trajectory that, according to the heuristic motion planning system, provides 
the shortest path to the target. 
This implies that a potential point of improvement for reducing the mission time of \gls{htmpc} 
is to include the mission time as an additional term in the objective function of TMPC.%

Based on Table~\ref{tab:cn_results}, for \textbf{case~2}, similarly to \gls{apf}, \gls{hlrrt} failed to show any success in reaching the target point.
In particular in setup~2, none of the failures of \gls{hlrrt} was due to colliding with any obstacles, but was mainly because, even with the larger computational budget, the algorithm 
failed to find any feasible paths.
From Figure~\ref{fig:cn_rrt_path}, \gls{hlrrt} explores a variety of options, which are quickly dismissed, because of the lack of dynamic prediction for the moving obstacles. 
The main reason for the back-and-forth motions is that \gls{hlrrt} followed a path for a while, 
which turned out to be infeasible later.%

For \gls{htmpc} (see right-hand side Table~\ref{tab:cn_results}) three failures for setup 2 occurred, where all relate to the challenging scenario illustrated in Figure~\ref{fig:crushing}.
The reason for more crashes in setup 1 for \gls{htmpc} compared to setup 2 was that  
the optimization solver reached the maximum number of iterations without finding a safe trajectory. 
In order to investigate alternative solutions, 
we considered an even larger computational budget, as well as a larger prediction horizon, which turned out to help avoiding crashes into the obstacles.%

\section*{Conclusions and topics for future research}
\label{ch:8cr}

In this paper, we proposed a novel control architecture for motion planning and reference tracking 
of autonomous robots in dynamic cluttered environments, based on a modified version of a heuristic motion planning method \cite{Jamshidnejad} and robust tube-based MPC.
In a case study, we compared our proposed approach to two state-of-the-art methods and showed, especially for complex scenarios, to have similar or significantly higher success rates and shorter path lengths with the proposed control architecture, while producing collision-free trajectories despite multiple moving obstacles.%

In the future, the proposed control architecture will be validated for different robot models, and for more variations in the motion of the obstacles, where proper filters should be merged into the control architecture to estimate the motion of these 
obstacles in real time. 
This control architecture should further be extended to multi-robot systems, 
with potentially heterogeneous characteristics.
Finally, real-life experiments should be performed to validate the effectiveness of the control architecture beyond computer-based simulations.%

\hspace{2ex}

\section*{Data Availability}

The data and parameters used in the simulations are publicly available in the 4TU.ResearchData repository: Baglioni, M. Tables with parameters values underlying the publication: Enabling robots to autonomously search dynamic cluttered
post-disaster environments. https://doi.org/10.4121/aa7528da-0986-453c-b196-4277a2db4daa (2024).

\small{\bibliography{Reference}}

\begin{thebibliography}{10}
\urlstyle{rm}
\expandafter\ifx\csname url\endcsname\relax
  \def\url#1{\texttt{#1}}\fi
\expandafter\ifx\csname urlprefix\endcsname\relax\def\urlprefix{URL }\fi
\expandafter\ifx\csname doiprefix\endcsname\relax\def\doiprefix{DOI: }\fi
\providecommand{\bibinfo}[2]{#2}
\providecommand{\eprint}[2][]{\url{#2}}

\bibitem{murphy2014disaster}
\bibinfo{author}{Murphy, R.~R.}
\newblock \emph{\bibinfo{title}{Disaster Robotics}} (\bibinfo{publisher}{MIT press}, \bibinfo{year}{2014}).

\bibitem{LiuNejat}
\bibinfo{author}{Liu, Y.} \& \bibinfo{author}{Nejat, G.}
\newblock \bibinfo{journal}{\bibinfo{title}{Robotic urban search and rescue: A survey from the control perspective}}.
\newblock {\emph{\JournalTitle{J. Intell. Robotics Syst.}}} \textbf{\bibinfo{volume}{72}}, \bibinfo{pages}{147–165}, \doiprefix\url{10.1007/s10846-013-9822-x} (\bibinfo{year}{2013}).

\bibitem{rajan2021disaster}
\bibinfo{author}{Rajan, J.}, \bibinfo{author}{Shriwastav, S.}, \bibinfo{author}{Kashyap, A.}, \bibinfo{author}{Ratnoo, A.} \& \bibinfo{author}{Ghose, D.}
\newblock \bibinfo{title}{Disaster management using unmanned aerial vehicles}.
\newblock In \emph{\bibinfo{booktitle}{Unmanned Aerial Systems}}, \bibinfo{pages}{129--155} (\bibinfo{publisher}{Elsevier}, \bibinfo{year}{2021}).

\bibitem{deKoning2023}
\bibinfo{author}{de~Koning, C.} \& \bibinfo{author}{Jamshidnejad, A.}
\newblock \bibinfo{journal}{\bibinfo{title}{Hierarchical integration of model predictive and fuzzy logic control for combined coverage and target-oriented search-and-rescue via robots with imperfect sensors}}.
\newblock {\emph{\JournalTitle{Journal of Intelligent \& Robotic Systems}}} \textbf{\bibinfo{volume}{107}}, \bibinfo{pages}{40} (\bibinfo{year}{2023}).

\bibitem{motionplanning:dynamic}
\bibinfo{author}{Mohanan, M.} \& \bibinfo{author}{Salgoankar, A.}
\newblock \bibinfo{journal}{\bibinfo{title}{A survey of robotic motion planning in dynamic environments}}.
\newblock {\emph{\JournalTitle{Robotics and Autonomous Systems}}} \textbf{\bibinfo{volume}{100}}, \bibinfo{pages}{171--185}, \doiprefix\url{https://doi.org/10.1016/j.robot.2017.10.011} (\bibinfo{year}{2018}).

\bibitem{motionplan:realtime}
\bibinfo{author}{Katrakazas, C.}, \bibinfo{author}{Quddus, M.}, \bibinfo{author}{Chen, W.-H.} \& \bibinfo{author}{Deka, L.}
\newblock \bibinfo{journal}{\bibinfo{title}{Real-time motion planning methods for autonomous on-road driving: State-of-the-art and future research directions}}.
\newblock {\emph{\JournalTitle{Transportation Research Part C: Emerging Technologies}}} \textbf{\bibinfo{volume}{60}}, \bibinfo{pages}{416--442}, \doiprefix\url{https://doi.org/10.1016/j.trc.2015.09.011} (\bibinfo{year}{2015}).

\bibitem{pandey2017mobile}
\bibinfo{author}{Pandey, A.}, \bibinfo{author}{Pandey, S.} \& \bibinfo{author}{Parhi, D.}
\newblock \bibinfo{journal}{\bibinfo{title}{Mobile robot navigation and obstacle avoidance techniques: A review}}.
\newblock {\emph{\JournalTitle{Int Rob Auto J}}} \textbf{\bibinfo{volume}{2}}, \bibinfo{pages}{00022} (\bibinfo{year}{2017}).

\bibitem{ohki2010collision}
\bibinfo{author}{Ohki, T.}, \bibinfo{author}{Nagatani, K.} \& \bibinfo{author}{Yoshida, K.}
\newblock \bibinfo{title}{Collision avoidance method for mobile robot considering motion and personal spaces of evacuees}.
\newblock In \emph{\bibinfo{booktitle}{2010 IEEE/RSJ International Conference on Intelligent Robots and Systems}}, \bibinfo{pages}{1819--1824} (\bibinfo{organization}{IEEE}, \bibinfo{year}{2010}).

\bibitem{liu2024design}
\bibinfo{author}{Liu, Z.}, \bibinfo{author}{Li, M.}, \bibinfo{author}{Fu, D.} \& \bibinfo{author}{Zhang, S.}
\newblock \bibinfo{journal}{\bibinfo{title}{Design of intelligent controller for obstacle avoidance and navigation of electric patrol mobile robot based on {PLC}}}.
\newblock {\emph{\JournalTitle{Scientific Reports}}} \textbf{\bibinfo{volume}{14}}, \bibinfo{pages}{13476} (\bibinfo{year}{2024}).

\bibitem{Jamshidnejad}
\bibinfo{author}{Jamshidnejad, A.} \& \bibinfo{author}{Frazzoli, E.}
\newblock \bibinfo{title}{Adaptive optimal receding-horizon robot navigation via short-term policy development}.
\newblock In \emph{\bibinfo{booktitle}{15th International Conference on Control, Automation, Robotics and Vision, {ICARCV} 2018, Singapore, November 18-21, 2018}}, \bibinfo{pages}{21--28}, \doiprefix\url{10.1109/ICARCV.2018.8581157} (\bibinfo{publisher}{{IEEE}}, \bibinfo{year}{2018}).

\bibitem{Surma2024}
\bibinfo{author}{Surma, F.} \& \bibinfo{author}{Jamshidnejad, A.}
\newblock \bibinfo{journal}{\bibinfo{title}{State-dependent dynamic tube {MPC}: A novel tube {MPC} method with a fuzzy model of disturbances}}.
\newblock {\emph{\JournalTitle{International Journal of Robust and Nonlinear Control}}}  (\bibinfo{year}{2024}).

\bibitem{hoy2015algorithms}
\bibinfo{author}{Hoy, M.}, \bibinfo{author}{Matveev, A.~S.} \& \bibinfo{author}{Savkin, A.~V.}
\newblock \bibinfo{journal}{\bibinfo{title}{Algorithms for collision-free navigation of mobile robots in complex cluttered environments: A survey}}.
\newblock {\emph{\JournalTitle{Robotica}}} \textbf{\bibinfo{volume}{33}}, \bibinfo{pages}{463--497} (\bibinfo{year}{2015}).

\bibitem{sar:review121}
\bibinfo{author}{Grogan, S.}, \bibinfo{author}{Pellerin, R.} \& \bibinfo{author}{Gamache, M.}
\newblock \bibinfo{journal}{\bibinfo{title}{The use of unmanned aerial vehicles and drones in search and rescue operations--a survey}}.
\newblock {\emph{\JournalTitle{Proceedings of the PROLOG}}}  (\bibinfo{year}{2018}).

\bibitem{nagasawa_et_al_2021}
\bibinfo{author}{Nagasawa, R.}, \bibinfo{author}{Mas, E.}, \bibinfo{author}{Moya, L.} \& \bibinfo{author}{Koshimura, S.}
\newblock \bibinfo{journal}{\bibinfo{title}{Model-based analysis of multi-{UAV} path planning for surveying postdisaster building damage}}.
\newblock {\emph{\JournalTitle{Scientific Reports}}} \textbf{\bibinfo{volume}{11}}, \bibinfo{pages}{18588} (\bibinfo{year}{2021}).

\bibitem{hashimoto_et_al_2022}
\bibinfo{author}{Hashimoto, A.}, \bibinfo{author}{Heintzman, L.}, \bibinfo{author}{Koester, R.} \& \bibinfo{author}{Abaid, N.}
\newblock \bibinfo{journal}{\bibinfo{title}{An agent-based model reveals lost person behavior based on data from wilderness search and rescue}}.
\newblock {\emph{\JournalTitle{Scientific Reports}}} \textbf{\bibinfo{volume}{12}}, \bibinfo{pages}{5873} (\bibinfo{year}{2022}).

\bibitem{serra_et_al_2020}
\bibinfo{author}{Serra, M.} \emph{et~al.}
\newblock \bibinfo{journal}{\bibinfo{title}{Search and rescue at sea aided by hidden flow structures}}.
\newblock {\emph{\JournalTitle{Nature Communications}}} \textbf{\bibinfo{volume}{11}}, \bibinfo{pages}{2525} (\bibinfo{year}{2020}).

\bibitem{kruijff_et_al_2012}
\bibinfo{author}{Kruijff, G.-J.~M.} \emph{et~al.}
\newblock \bibinfo{title}{Rescue robots at earthquake-hit {M}irandola, {I}taly: A field report}.
\newblock In \emph{\bibinfo{booktitle}{2012 IEEE International Symposium on Safety, Security, and Rescue Robotics (SSRR)}}, \bibinfo{pages}{1--8} (\bibinfo{organization}{IEEE}, \bibinfo{year}{2012}).

\bibitem{schedl_et_al_2020}
\bibinfo{author}{Schedl, D.~C.}, \bibinfo{author}{Kurmi, I.} \& \bibinfo{author}{Bimber, O.}
\newblock \bibinfo{journal}{\bibinfo{title}{Search and rescue with airborne optical sectioning}}.
\newblock {\emph{\JournalTitle{Nature Machine Intelligence}}} \textbf{\bibinfo{volume}{2}}, \bibinfo{pages}{783--790} (\bibinfo{year}{2020}).

\bibitem{fattah2016r3diver}
\bibinfo{author}{Fattah, S.} \emph{et~al.}
\newblock \bibinfo{title}{R3{D}iver: Remote robotic rescue diver for rapid underwater search and rescue operation}.
\newblock In \emph{\bibinfo{booktitle}{2016 IEEE Region 10 Conference (TENCON)}}, \bibinfo{pages}{3280--3283} (\bibinfo{organization}{IEEE}, \bibinfo{year}{2016}).

\bibitem{colas_et_al_2013}
\bibinfo{author}{Colas, F.}, \bibinfo{author}{Mahesh, S.}, \bibinfo{author}{Pomerleau, F.}, \bibinfo{author}{Liu, M.} \& \bibinfo{author}{Siegwart, R.}
\newblock \bibinfo{title}{3{D} path planning and execution for search and rescue ground robots}.
\newblock In \emph{\bibinfo{booktitle}{2013 IEEE/RSJ International Conference on Intelligent Robots and Systems}}, \bibinfo{pages}{722--727} (\bibinfo{organization}{IEEE}, \bibinfo{year}{2013}).

\bibitem{arnold2020heterogeneous}
\bibinfo{author}{Arnold, R.}, \bibinfo{author}{Jablonski, J.}, \bibinfo{author}{Abruzzo, B.} \& \bibinfo{author}{Mezzacappa, E.}
\newblock \bibinfo{title}{Heterogeneous {UAV} multi-role swarming behaviors for search and rescue}.
\newblock In \emph{\bibinfo{booktitle}{2020 IEEE Conference on Cognitive and Computational Aspects of Situation Management (CogSIMA)}}, \bibinfo{pages}{122--128} (\bibinfo{organization}{IEEE}, \bibinfo{year}{2020}).

\bibitem{san2018intelligent}
\bibinfo{author}{San~Juan, V.}, \bibinfo{author}{Santos, M.} \& \bibinfo{author}{And{\'u}jar, J.~M.}
\newblock \bibinfo{journal}{\bibinfo{title}{Intelligent {UAV} map generation and discrete path planning for search and rescue operations}}.
\newblock {\emph{\JournalTitle{Complexity}}} \textbf{\bibinfo{volume}{2018}}, \bibinfo{pages}{6879419} (\bibinfo{year}{2018}).

\bibitem{loukas2008connecting}
\bibinfo{author}{Loukas, G.} \& \bibinfo{author}{Timotheou, S.}
\newblock \bibinfo{title}{Connecting trapped civilians to a wireless ad hoc network of emergency response robots}.
\newblock In \emph{\bibinfo{booktitle}{2008 11th IEEE Singapore International Conference on Communication Systems}}, \bibinfo{pages}{599--603} (\bibinfo{organization}{IEEE}, \bibinfo{year}{2008}).

\bibitem{davids2002urban}
\bibinfo{author}{Davids, A.}
\newblock \bibinfo{journal}{\bibinfo{title}{Urban search and rescue robots: From tragedy to technology}}.
\newblock {\emph{\JournalTitle{IEEE Intelligent Systems}}} \textbf{\bibinfo{volume}{17}}, \bibinfo{pages}{81--83} (\bibinfo{year}{2002}).

\bibitem{bogue2019disaster}
\bibinfo{author}{Bogue, R.}
\newblock \bibinfo{journal}{\bibinfo{title}{Disaster relief, and search and rescue robots: The way forward}}.
\newblock {\emph{\JournalTitle{Industrial Robot: the International Journal of Robotics Research and Application}}} \textbf{\bibinfo{volume}{46}}, \bibinfo{pages}{181--187} (\bibinfo{year}{2019}).

\bibitem{stecz2020uav}
\bibinfo{author}{Stecz, W.} \& \bibinfo{author}{Gromada, K.}
\newblock \bibinfo{journal}{\bibinfo{title}{{UAV} mission planning with {SAR} application}}.
\newblock {\emph{\JournalTitle{Sensors}}} \textbf{\bibinfo{volume}{20}}, \bibinfo{pages}{1080} (\bibinfo{year}{2020}).

\bibitem{berger2015innovative}
\bibinfo{author}{Berger, J.} \& \bibinfo{author}{Lo, N.}
\newblock \bibinfo{journal}{\bibinfo{title}{An innovative multi-agent search-and-rescue path planning approach}}.
\newblock {\emph{\JournalTitle{Computers \& Operations Research}}} \textbf{\bibinfo{volume}{53}}, \bibinfo{pages}{24--31} (\bibinfo{year}{2015}).

\bibitem{de2019autonomous}
\bibinfo{author}{de~Alcantara~Andrade, F.~A.} \emph{et~al.}
\newblock \bibinfo{journal}{\bibinfo{title}{Autonomous unmanned aerial vehicles in search and rescue missions using real-time cooperative model predictive control}}.
\newblock {\emph{\JournalTitle{Sensors}}} \textbf{\bibinfo{volume}{19}}, \bibinfo{pages}{4067} (\bibinfo{year}{2019}).

\bibitem{baglioni_jamshidnejad_2024}
\bibinfo{author}{Baglioni, M.} \& \bibinfo{author}{Jamshidnejad, A.}
\newblock \bibinfo{journal}{\bibinfo{title}{A novel {MPC} formulation for dynamic target tracking with increased area coverage for search-and-rescue robots}}.
\newblock {\emph{\JournalTitle{Journal of Intelligent \& Robotic Systems}}} \textbf{\bibinfo{volume}{110}}, \bibinfo{pages}{140} (\bibinfo{year}{2024}).

\bibitem{berger_lo_2015}
\bibinfo{author}{Berger, J.} \& \bibinfo{author}{Lo, N.}
\newblock \bibinfo{journal}{\bibinfo{title}{An innovative multi-agent search-and-rescue path planning approach}}.
\newblock {\emph{\JournalTitle{Computers \& Operations Research}}} \textbf{\bibinfo{volume}{53}}, \bibinfo{pages}{24--31} (\bibinfo{year}{2015}).

\bibitem{hoy_et_al_2012}
\bibinfo{author}{Hoy, M.}, \bibinfo{author}{Matveev, A.~S.} \& \bibinfo{author}{Savkin, A.~V.}
\newblock \bibinfo{journal}{\bibinfo{title}{Collision free cooperative navigation of multiple wheeled robots in unknown cluttered environments}}.
\newblock {\emph{\JournalTitle{Robotics and Autonomous Systems}}} \textbf{\bibinfo{volume}{60}}, \bibinfo{pages}{1253--1266} (\bibinfo{year}{2012}).

\bibitem{farrokhsiar_et_al_2013}
\bibinfo{author}{Farrokhsiar, M.}, \bibinfo{author}{Pavlik, G.} \& \bibinfo{author}{Najjaran, H.}
\newblock \bibinfo{journal}{\bibinfo{title}{An integrated robust probing motion planning and control scheme: A tube-based {MPC} approach}}.
\newblock {\emph{\JournalTitle{Robotics and Autonomous Systems}}} \textbf{\bibinfo{volume}{61}}, \bibinfo{pages}{1379--1391} (\bibinfo{year}{2013}).

\bibitem{nattero_et_al_2014}
\bibinfo{author}{Nattero, C.}, \bibinfo{author}{Recchiuto, C.~T.}, \bibinfo{author}{Sgorbissa, A.} \& \bibinfo{author}{Wanderlingh, F.}
\newblock \bibinfo{title}{Coverage algorithms for search and rescue with {UAV} drones}.
\newblock In \emph{\bibinfo{booktitle}{Artificial Intelligence, Workshop of the XIII AI* IA Symposium on}}, vol.~\bibinfo{volume}{12} (\bibinfo{year}{2014}).

\bibitem{galceran_carreras_2013}
\bibinfo{author}{Galceran, E.} \& \bibinfo{author}{Carreras, M.}
\newblock \bibinfo{journal}{\bibinfo{title}{A survey on coverage path planning for robotics}}.
\newblock {\emph{\JournalTitle{Robotics and Autonomous Systems}}} \textbf{\bibinfo{volume}{61}}, \bibinfo{pages}{1258--1276} (\bibinfo{year}{2013}).

\bibitem{brooks2009randomised}
\bibinfo{author}{Brooks, A.}, \bibinfo{author}{Kaupp, T.} \& \bibinfo{author}{Makarenko, A.}
\newblock \bibinfo{title}{Randomised {MPC}-based motion-planning for mobile robot obstacle avoidance}.
\newblock In \emph{\bibinfo{booktitle}{2009 IEEE International Conference on Robotics and Automation}}, \bibinfo{pages}{3962--3967} (\bibinfo{organization}{IEEE}, \bibinfo{year}{2009}).

\bibitem{paez_et_al_2021}
\bibinfo{author}{Paez, D.}, \bibinfo{author}{Romero, J.~P.}, \bibinfo{author}{Noriega, B.}, \bibinfo{author}{Cardona, G.~A.} \& \bibinfo{author}{Calderon, J.~M.}
\newblock \bibinfo{journal}{\bibinfo{title}{Distributed particle swarm optimization for multi-robot system in search and rescue operations}}.
\newblock {\emph{\JournalTitle{IFAC-PapersOnLine}}} \textbf{\bibinfo{volume}{54}}, \bibinfo{pages}{1--6} (\bibinfo{year}{2021}).

\bibitem{niroui_et_al_2019}
\bibinfo{author}{Niroui, F.}, \bibinfo{author}{Zhang, K.}, \bibinfo{author}{Kashino, Z.} \& \bibinfo{author}{Nejat, G.}
\newblock \bibinfo{journal}{\bibinfo{title}{Deep reinforcement learning robot for search and rescue applications: Exploration in unknown cluttered environments}}.
\newblock {\emph{\JournalTitle{IEEE Robotics and Automation Letters}}} \textbf{\bibinfo{volume}{4}}, \bibinfo{pages}{610--617} (\bibinfo{year}{2019}).

\bibitem{mohseni_et_al_2017}
\bibinfo{author}{Mohseni, F.}, \bibinfo{author}{Doustmohammadi, A.} \& \bibinfo{author}{Menhaj, M.~B.}
\newblock \bibinfo{journal}{\bibinfo{title}{Distributed model predictive coverage control for decoupled mobile robots}}.
\newblock {\emph{\JournalTitle{Robotica}}} \textbf{\bibinfo{volume}{35}}, \bibinfo{pages}{922--941} (\bibinfo{year}{2017}).

\bibitem{ibrahim_et_al_2019}
\bibinfo{author}{Ibrahim, M.}, \bibinfo{author}{Matschek, J.}, \bibinfo{author}{Morabito, B.} \& \bibinfo{author}{Findeisen, R.}
\newblock \bibinfo{journal}{\bibinfo{title}{Hierarchical model predictive control for autonomous vehicle area coverage}}.
\newblock {\emph{\JournalTitle{IFAC-PapersOnLine}}} \textbf{\bibinfo{volume}{52}}, \bibinfo{pages}{79--84} (\bibinfo{year}{2019}).

\bibitem{BemporadMorari}
\bibinfo{author}{Bemporad, A.} \& \bibinfo{author}{Morari, M.}
\newblock \bibinfo{title}{Robust model predictive control: A survey}.
\newblock In \emph{\bibinfo{booktitle}{Robustness in Identification and Control}}, \bibinfo{pages}{207--226} (\bibinfo{publisher}{Springer}, \bibinfo{year}{1999}).

\bibitem{langson_chryssochoos_rakovic_mayne_2004}
\bibinfo{author}{Langson, W.}, \bibinfo{author}{Chryssochoos, I.}, \bibinfo{author}{Rakovi$\acute{c}$, S.~V.} \& \bibinfo{author}{Mayne, D.~Q.}
\newblock \bibinfo{journal}{\bibinfo{title}{Robust model predictive control using tubes}}.
\newblock {\emph{\JournalTitle{Automatica}}} \textbf{\bibinfo{volume}{40}}, \bibinfo{pages}{125--133} (\bibinfo{year}{2004}).

\bibitem{liu2019recursive}
\bibinfo{author}{Liu, Z.} \& \bibinfo{author}{Stursberg, O.}
\newblock \bibinfo{title}{Recursive feasibility and stability of {MPC} with time-varying and uncertain state constraints}.
\newblock In \emph{\bibinfo{booktitle}{2019 18th European Control Conference (ECC)}}, \bibinfo{pages}{1766--1771} (\bibinfo{organization}{IEEE}, \bibinfo{year}{2019}).

\bibitem{mayne2011tube}
\bibinfo{author}{Mayne, D.~Q.}, \bibinfo{author}{Kerrigan, E.~C.}, \bibinfo{author}{Van~Wyk, E.} \& \bibinfo{author}{Falugi, P.}
\newblock \bibinfo{journal}{\bibinfo{title}{Tube-based robust nonlinear model predictive control}}.
\newblock {\emph{\JournalTitle{International Journal of Robust and Nonlinear Control}}} \textbf{\bibinfo{volume}{21}}, \bibinfo{pages}{1341--1353} (\bibinfo{year}{2011}).

\bibitem{wang_tan_nejat_2024}
\bibinfo{author}{Wang, H.}, \bibinfo{author}{Tan, A.~H.} \& \bibinfo{author}{Nejat, G.}
\newblock \bibinfo{journal}{\bibinfo{title}{Nav{F}ormer: A transformer architecture for robot target-driven navigation in unknown and dynamic environments}}.
\newblock {\emph{\JournalTitle{IEEE Robotics and Automation Letters}}}  (\bibinfo{year}{2024}).

\bibitem{guo2009combination}
\bibinfo{author}{Guo, Y.}, \bibinfo{author}{Song, A.}, \bibinfo{author}{Bao, J.}, \bibinfo{author}{Hongru, T.} \& \bibinfo{author}{Cui, J.}
\newblock \bibinfo{journal}{\bibinfo{title}{A combination of terrain prediction and correction for search and rescue robot autonomous navigation}}.
\newblock {\emph{\JournalTitle{International Journal of Advanced Robotic Systems}}} \textbf{\bibinfo{volume}{6}}, \bibinfo{pages}{24} (\bibinfo{year}{2009}).

\bibitem{devo_et_al_2020}
\bibinfo{author}{Devo, A.}, \bibinfo{author}{Mezzetti, G.}, \bibinfo{author}{Costante, G.}, \bibinfo{author}{Fravolini, M.~L.} \& \bibinfo{author}{Valigi, P.}
\newblock \bibinfo{journal}{\bibinfo{title}{Towards generalization in target-driven visual navigation by using deep reinforcement learning}}.
\newblock {\emph{\JournalTitle{IEEE Transactions on Robotics}}} \textbf{\bibinfo{volume}{36}}, \bibinfo{pages}{1546--1561} (\bibinfo{year}{2020}).

\bibitem{sani_et_al_2021}
\bibinfo{author}{Sani, M.}, \bibinfo{author}{Robu, B.} \& \bibinfo{author}{Hably, A.}
\newblock \bibinfo{title}{Pursuit-evasion games based on game-theoretic and model predictive control algorithms}.
\newblock In \emph{\bibinfo{booktitle}{2021 International Conference on Control, Automation and Diagnosis (ICCAD)}}, \bibinfo{pages}{1--6} (\bibinfo{organization}{IEEE}, \bibinfo{year}{2021}).

\bibitem{dhaouadi2013dynamic}
\bibinfo{author}{Dhaouadi, R.} \& \bibinfo{author}{Hatab, A.~A.}
\newblock \bibinfo{journal}{\bibinfo{title}{Dynamic modelling of differential-drive mobile robots using {L}agrange and {N}ewton-{E}uler methodologies: A unified framework}}.
\newblock {\emph{\JournalTitle{Advances in Robotics \& Automation}}} \textbf{\bibinfo{volume}{2}}, \bibinfo{pages}{1--7} (\bibinfo{year}{2013}).

\bibitem{hierarchicalmpc}
\bibinfo{author}{Scattolini, R.}
\newblock \bibinfo{journal}{\bibinfo{title}{Architectures for distributed and hierarchical model predictive control - a review}}.
\newblock {\emph{\JournalTitle{Journal of Process Control}}} \textbf{\bibinfo{volume}{19}}, \bibinfo{pages}{723--731} (\bibinfo{year}{2009}).

\bibitem{basescu2020direct}
\bibinfo{author}{Basescu, M.} \& \bibinfo{author}{Moore, J.}
\newblock \bibinfo{title}{Direct {NMPC} for post-stall motion planning with fixed-wing {UAV}s}.
\newblock In \emph{\bibinfo{booktitle}{2020 IEEE International Conference on Robotics and Automation (ICRA)}}, \bibinfo{pages}{9592--9598} (\bibinfo{organization}{IEEE}, \bibinfo{year}{2020}).

\bibitem{esteves_et_al_2024}
\bibinfo{author}{Esteves~Henriques, B.}, \bibinfo{author}{Baglioni, M.} \& \bibinfo{author}{Jamshidnejad, A.}
\newblock \bibinfo{journal}{\bibinfo{title}{Camera-based mapping in search-and-rescue via flying and ground robot teams}}.
\newblock {\emph{\JournalTitle{Machine Vision and Applications}}} \textbf{\bibinfo{volume}{35}}, \bibinfo{pages}{117} (\bibinfo{year}{2024}).

\bibitem{mitchell_1998}
\bibinfo{author}{Mitchell, M.}
\newblock \emph{\bibinfo{title}{An Introduction to Genetic Algorithms}} (\bibinfo{publisher}{MIT press}, \bibinfo{year}{1998}).

\bibitem{torczon_1997}
\bibinfo{author}{Torczon, V.}
\newblock \bibinfo{journal}{\bibinfo{title}{On the convergence of pattern search algorithms}}.
\newblock {\emph{\JournalTitle{SIAM Journal on Optimization}}} \textbf{\bibinfo{volume}{7}}, \bibinfo{pages}{1--25} (\bibinfo{year}{1997}).

\bibitem{chen2019horizon}
\bibinfo{author}{Chen, Y.}, \bibinfo{author}{He, Z.} \& \bibinfo{author}{Li, S.}
\newblock \bibinfo{journal}{\bibinfo{title}{Horizon-based lazy optimal {RRT} for fast, efficient replanning in dynamic environment}}.
\newblock {\emph{\JournalTitle{Autonomous Robots}}} \textbf{\bibinfo{volume}{43}}, \bibinfo{pages}{2271--2292} (\bibinfo{year}{2019}).

\bibitem{lyu2019colregs}
\bibinfo{author}{Lyu, H.} \& \bibinfo{author}{Yin, Y.}
\newblock \bibinfo{journal}{\bibinfo{title}{{COLREGS}-constrained real-time path planning for autonomous ships using modified artificial potential fields}}.
\newblock {\emph{\JournalTitle{The Journal of Navigation}}} \textbf{\bibinfo{volume}{72}}, \bibinfo{pages}{588--608} (\bibinfo{year}{2019}).

\bibitem{baglioni_dataset_2024}
\bibinfo{author}{Baglioni, M.}
\newblock \bibinfo{title}{Tables with parameters values underlying the publication: Enabling robots to autonomously search dynamic cluttered post-disaster environments}.
\newblock \bibinfo{howpublished}{\url{https://doi.org/10.4121/aa7528da-0986-453c-b196-4277a2db4daa}} (\bibinfo{year}{2024}).

\bibitem{bruce2002real}
\bibinfo{author}{Bruce, J.} \& \bibinfo{author}{Veloso, M.~M.}
\newblock \bibinfo{title}{Real-time randomized path planning for robot navigation}.
\newblock In \emph{\bibinfo{booktitle}{Robot Soccer World Cup}}, \bibinfo{pages}{288--295} (\bibinfo{organization}{Springer}, \bibinfo{year}{2002}).

\bibitem{yang2012robust}
\bibinfo{author}{Yang, M.-S.}, \bibinfo{author}{Lai, C.-Y.} \& \bibinfo{author}{Lin, C.-Y.}
\newblock \bibinfo{journal}{\bibinfo{title}{A robust {EM} clustering algorithm for {G}aussian mixture models}}.
\newblock {\emph{\JournalTitle{Pattern Recognition}}} \textbf{\bibinfo{volume}{45}}, \bibinfo{pages}{3950--3961} (\bibinfo{year}{2012}).

\end{thebibliography}

\section*{Acknowledgements}

This research has been supported jointly by the NWO Talent Program Veni project ``Autonomous drones flocking for search-and-rescue'' (18120), which has been financed by the Netherlands Organisation for Scientific Research (NWO) and by the TU Delft AI Labs \& Talent programme.

\section*{Author contributions}

K. Rado performed the conceptualization, designed and performed the simulations, contributed to methodology, analysis and validity assessment of the results, and wrote the first draft of the paper. 
M. Baglioni contributed to methodology, analysis and validity assessment of the results, supervision, and reviewing the final draft of the paper.
A. Jamshidnejad contributed to conceptualization, methodology, analysis and validity assessment of the results, editing the final draft of the paper, project administration and supervision, and funding acquisition.

\section*{Competing interests}

The authors declare that they have no competing interests.

\end{document}